\documentclass[preprint, prX]{revtex4}

\usepackage{amsmath}    
\usepackage{graphicx}   
\usepackage{verbatim}   
\usepackage{color}      
\usepackage{subfigure}  
\usepackage{hyperref}   
\usepackage{amssymb}    
\usepackage{multirow}
\usepackage{epsfig}
\usepackage{graphics,graphicx}
\usepackage{setspace}
\usepackage{url}
\usepackage{algorithm,algorithmic}
\usepackage{sidecap}

\begin{document}

\begin{abstract}
This paper proposes a hybrid self-adaptive evolutionary algorithm for graph coloring that is hybridized with the following novel
elements: heuristic genotype-phenotype mapping, a swap local search heuristic, and a neutral survivor selection operator. This
algorithm was compared with the evolutionary algorithm with the SAW method of Eiben et al., the Tabucol algorithm of Hertz and de Werra, and the hybrid evolutionary algorithm of Galinier and Hao. The performance of these algorithms were tested on a test suite consisting
of randomly generated 3-colorable graphs of various structural features, such as graph size, type, edge density, and variability in
sizes of color classes. Furthermore, the test graphs were generated including the phase transition where the graphs are hard to color.
The purpose of the extensive experimental work was threefold: to investigate the behavior of the tested algorithms in
the phase transition, to identify what impact hybridization with the DSatur traditional heuristic has on the evolutionary algorithm, and to show how graph structural features influence the performance of the graph-coloring algorithms. The results indicate that the performance of the hybrid self-adaptive evolutionary algorithm is comparable with, or better than, the performance of the hybrid evolutionary algorithm which is one of the best graph-coloring algorithms today. Moreover, the fact that all the considered algorithms performed poorly on flat graphs confirms that this type of graphs is really the hardest to color.

Original paper: \url{http://dx.doi.org/10.1007/s10589-012-9496-5}

\end{abstract}

\title{Graph 3-coloring with a hybrid self-adaptive evolutionary algorithm}

\author{Iztok Fister}
\altaffiliation{University of Maribor, Faculty of electrical engineering and computer science
Smetanova 17, 2000 Maribor}
\email{iztok.fister@uni-mb.si}

\author{Marjan Mernik}
\altaffiliation{University of Maribor, Faculty of electrical engineering and computer science
Smetanova 17, 2000 Maribor}
\email{marjan.mernik@uni-mb.si}

\author{Bogdan Filipi\v{c}}
\altaffiliation{Department of Intelligent Systems, Jo\v{z}ef Stefan Institute,
Jamova cesta 39, 1000 Ljubljana}
\email{bogdan.filipic@ijs.si}

\maketitle

\section{Introduction}
\label{intro}
The graph-coloring problem (GCP) is a well-known combinatorial optimization problem that has many practical applications. For example,
it can be applied in register allocation in compilers~\cite{Chow:1990,deWerra:1999}, timetabling~\cite{Burke:2007,deWerra:1985},
frequency assignment in cellular networks~\cite{Gamst:1986,Mabed:2010,Smith:1998}, scheduling~\cite{Gamache:2007,Leighton:1979}, printed circuit board testing~\cite{Garey:1976}, and manufacturing~\cite{Glass:2002}. In addition to its practical relevance, the problem
represents a test bed for new algorithms primarily because of its simple definition.

The problem can be defined informally as follows: In a given undirected graph $G=(V,E)$, where $V$ denotes a finite set of vertices and
$E$ a finite set of unordered pairs of vertices called edges, the vertices have to be colored using $k$ colors such that no two vertices connected with an edge are of the same color. Such a coloring, if it exists, is named the \textit{proper} $k$-coloring. The minimum number of colors $k$ necessary to color the graph $G$ is called the \textit{chromatic number} $\chi$~\cite{Kubale:2004}. The graph is $k$-colorable if it has a $k$-coloring. Searching for a proper graph $k$-coloring is denoted as $k$-GCP. The decision form of this problem, where the question is whether a particular graph is $k$-colorable, is $\mathit{NP}$-complete~\cite{Garey:1979}, while the problem of finding the chromatic number of graph $G$ is $\mathit{NP}$-hard \cite{Bondy:2008}.

The first algorithms for graph coloring tried to solve this problem exactly~\cite{Brown:1972}, i.e., by enumerating all possible
orderings of the vertices. However, these algorithms were too time consuming and infeasible for large graphs, therefore, many approaches that tackled the problem heuristically were proposed. They tried to find an approximate solution in reasonable time. The most natural
approach to coloring graph vertices heuristically was the greedy approach~\cite{Bondy:2008}. The best known heuristics of this type are
the \textit{largest saturation degree} heuristic DSatur~\cite{Brelaz:1979}, and the \textit{recursive largest first} heuristic
RLF~\cite{Leighton:1979}. Before coloring, DSatur orders graph vertices according to the \textit{saturation degree} $\rho_{v}$ that is
defined as the number of distinctly colored vertices adjacent to vertex $v$~\cite{Kubale:2004}. On the other hand, RLF divides the
uncolored graph into color classes that contain vertices colored with the same color. The algorithm colors the vertices one color
class at a time. The vertices from the uncolored subgraph are added to the current color class in turn so that the number of edges left
in the uncolored subgraph remains as few as possible.

Some of the most popular algorithms for solving $k$-GCP today are metaheuristics based on local
search~\cite{Blum:2011,Blum:2003}. One of the first metaheuristics was developed by Hertz and de
Werra~\cite{Hertz:1987} under the name Tabucol. This was the first application of the tabu search~\cite{Glover:1986} to graph coloring.
Tabucol first generates an initial random $k$-coloring, which typically contains a large number of conflicting edges. Then, the
heuristic iteratively looks for a single vertex that most decreases the number of conflicting edges when it is recolored with another
color, i.e., moved to another color class. A tabu list prevents the moves from cycling. Proper $k$-coloring may be
obtained after a finite number of iterations. Later, Tabucol was improved to more sophisticated graph-coloring
algorithms~\cite{Dorne:1998,Galinier:1999}.

On the other hand, other local search heuristics include simulated annealing~\cite{Chams:1987,Johnson:1991}, iterative local
search~\cite{Chiarandini:2002,Chiarandini:2007}, reactive partial tabu search~\cite{Blochliger:2008,Blochliger2003,Malaguti:2008},
variable neighborhood search~\cite{Avanthay:2003}, adaptive memory~\cite{Galinier:2008}, variable search space~\cite{Hertz:2008}, and
population-based methods~\cite{Dorne:1998,Fleurent:1996,Galinier:1999,Lu:2010}. One of the best po\-pu\-la\-tion-ba\-sed algorithms for
$k$-GCP, the Hybrid Evolutionary Algorithm (HEA) developed by Galiner and Hao~\cite{Galinier:1999} combines local search with the
partition-based crossover operator. The Tabucol metaheuristic is used as a local search operator. For a comprehensive survey of the
main methods, see, e.g.,~\cite{Galinier:2006,Malaguti:2009}.

In general, evolutionary algorithms~\cite{Michalewicz:1992} that operate on a population of solutions are good optimizers, but suffer
from a lack of constraint handling abilities. This lack arises from the fact that the variation operators, i.e., crossover and mutation, tend to violate the constraints~\cite{Eiben:2003}. As a result, offspring generated from parents by these operators can be infeasible.
Indirect constraint handling can be used to overcome this problem. In this case, constraints are transformed into optimization
objectives that can be expressed by an objective function. This function penalizes the candidate solutions violating the constraints. A solution to the problem is found when all constraints are satisfied. The value of the objective function in this case decreases to
zero.

GCP is a constraint optimization problem where two vertices connected with an edge cannot be colored with the same color. However,
the graph vertices have different numbers of edges. The number of edges incident to vertex $v$ is called the \textit{vertex degree}
$\mathit{deg}_{G}(v)$~\cite{Bondy:2008}. Regarding the vertex degrees, it may be harder to color vertices with higher rather than lower degree. This dependence can be expressed by an objective function defined as the sum of the weights assigned to the vertices violating the constraints. The higher the weight of a vertex, the more difficult the vertex is to color. The manner in which the weights are changed has a great influence on the performance of a graph-coloring algorithm. Weights can be changed deterministically, adaptively or self-adaptively~\cite{Eiben:1999}.

This paper focuses on graph 3-coloring (3-GCP) that is a special case of $k$-GCP, where $k=3$. It proposes a self-adaptive
evolutionary algorithm for 3-GCP hybridized with:
\begin{itemize}
  \item heuristic genotype-phenotype mapping (construction of solutions),
  \item a local search heuristic,
  \item a neutral survivor selection operator.
\end{itemize}

3-GCP is often the subject of research because of the lowest $k$ where $k$-GCP makes any sense. This work was motivated by
the paper of Eiben et al.~\cite{Eiben:1998}, which was our starting point for 3-GCP. Eiben et al. describe an evolutionary algorithm with Stepwise Adaptation of Weights (SAW), denoted as SAW-EA. Solutions in this evolutionary algorithm are represented as permutations of vertices and corresponding weights that denote the hardness of the vertices to be colored. This hardness is expressed by the penalty reward reflected in the objective function. In order to minimize this objective function, the graph-coloring algorithm (greedy algorithm) directs itself to first color the vertices with higher values of weights. Note that here the weights are modified adaptively. The results of the proposed self-adaptive evolutionary algorithm were compared with the results of three graph-coloring algorithms: SAW-EA~\cite{website:Hemert}, Tabucol~\cite{website:Chiarandini} and HEA~\cite{website:Chiarandini}. Extensive experiments using these algorithms were conducted on a collection of random 3-colorable graphs generated by the Culbersone graph generator~\cite{website:Culberson}. The test suite used in the experiments was the same as in~\cite{Eiben:1998} because the DIMACS challenge suite~\cite{Johnson:1996}, which was designed for testing graph $k$-coloring algorithms, does not contain 3-colorable graphs. Our test suite satisfies two conditions:
\begin{itemize}
  \item it comprises the instances of the random graphs in the phase transition~\cite{Turner:1988}, where the graphs are really hard to color,
  \item it enables the determination of how various structural features of the random graphs, e.g., the graph size and type, the edge density, and the variability in sizes of color classes, influence the performance of the tested graph-coloring algorithms.
\end{itemize}

In summary, the hybrid self-adaptive evolutionary algorithm incorporates three novelties: adaptation of weights representing the coefficients of the objective function in the DSatur algorithm, a new local search heuristic, and a new operator of the neutral survivor selection. Although this approach recalls SAW-EA, at least two differences can be exposed. The SAW method adapts the coefficients of the objective function in the greedy coloring heuristic (in contrast to the DSatur heuristic) and is adaptive (in contrast to
self-adaptation). Furthermore, while Igel \& Taussaint~\cite{Igel:2003} emphasize the significance of neutral mutation in evolution
strategies, the proposed approach exploits this finding explicitly by creating the neutral survivor selection operator. The contribution of this evolutionary algorithm  to the original DSatur algorithm was evaluated during the experimental work.

The structure of the rest of this paper is as follows. In Section 2 the problem of graph 3-coloring is formally defined. Section 3
describes the structure of the hybrid self-adaptive evolutionary algorithm in detail. Section 4 presents experiments and results  on various classes of random graphs. Section 5 concludes the paper by summarizing the work, the experimental results, and ideas
for further work.

\section{Graph 3-coloring}
\label{sec:1}

Let us assume an undirected graph $G=(V,E)$, where $V$ represents the finite set of vertices $v_{i} \in V$ and $E$ the finite set of
unordered pairs $e=\{v_{i},v_{j}\}$ named edges for $i=1 \ldots n\ \wedge\ j=1 \ldots n \wedge i\ \neq\ j$, where, $n=|V|$ denotes the number of vertices. A graph 3-coloring is a mapping $c:V \rightarrow C$, where $C=\{1,2,3\}$ denotes a set of 3-colors. In other words, one of the three colors is assigned to each vertex of the graph $G$. A proper 3-coloring is obtained if no two adjacent vertices are assigned the same color~\cite{Bondy:2008}.

3-GCP is a well-known \textit{constraint satisfaction problem} (CSP)~\cite{Eiben:2003} convenient for solving with evolutionary
algorithms. CSP can be represented as a pair $\langle S, \phi \rangle$, where $S=C^{n}$ with $C=\{1,2,3\}$ denoting a \textit{search
space}, in which all solutions $\textbf{s}= \langle s_{1}, \ldots ,s_{n} \rangle \in S$ are \textit{feasible}
and $\phi$ a \textit{feasibility condition}, i.e., a Boolean function on $S$, that is composed of constraints belonging to edges. That
is, for each edge $e \in E$ the corresponding constraint $b_{e}$ is defined by $b_{e}(\langle s_{1}, \ldots ,s_{n} \rangle)=true$ if and only if $e=\{v_{i},v_{j}\}$ and $s_{i} \neq s_{j}$. Let $B^{i}=\{b_{e}|e=\{v_{i},v_{j}\} \wedge j=1 \ldots m\}$ be the set of
constraints involving variable $v_{i}$ (edges connecting to node $i$). Then the feasibility condition is the conjunction of all
constraints $\phi (\textbf{s})= \wedge_{v \in V} B^{i}(\textbf{s})$. Note that the feasibility condition divides the search space into
\textit{feasible} and \textit{infeasible regions}.

Constraints are usually handled in evolutionary algorithms indirectly through a penalty function that transforms the CSP into the
\textit{free optimization problem} (FOP)~\cite{Eiben:2003}. Thus, those infeasible solutions that are far away from the feasible region
are punished with higher penalties. The penalty function for 3-GCP that can also be used as an objective function is

\begin{equation}
\label{eq:1}
 f(\textbf{s})= \displaystyle\sum_{i=1}^{n} \psi(\textbf{s},B^{i}),
\end{equation}
\noindent

\noindent where the function $\psi(\textbf{s},B^{i})$ is defined as

\begin{equation}
\label{eq:2}
\psi(\textbf{s},B^{i})=\left\{\begin{matrix}
 1 & \text{if } \textbf{s} \text{ violates at least one } b_{e} \in B^{i}\,, \\
 0 & \text{otherwise}\,.
\end{matrix}\right.
\end{equation}
					
The penalty function in Eq.~\ref{eq:1} transforms the constraint satisfaction problem into a free optimization problem such that for
each $\textbf{s} \in S$ we have $\phi(\textbf{s})=true$ if and only if $f(\textbf{s})=0$. Eq.~\ref{eq:2} represents the feasibility
condition and estimates the quality of candidate solution $\textbf{s}$. In fact, Eq.~\ref{eq:1} counts the number of vertices
that violate the constraints as expressed by Eq.~\ref{eq:2}.

\section{The hybrid self-adaptive evolutionary algorithm}
\label{sec:2}

The hybrid self-adaptive evolutionary algorithm (HSA-EA) for graph 3-coloring integrates concepts from various problem solving methods.
As a base, the self-adaptive evolution strategy~\cite{Baeck:1996} is used and then hybridized with heuristic genotype-phenotype
mapping, a local search heuristic, and the neutral survivor selection.

\subsection{The algorithms outline}
\label{ssec:3}

HSA-EA involves the following components and features:

\begin{itemize}
  \item representation of individuals,
  \item evaluation of objective function,
  \item population model,
  \item parent selection,
  \item mutation operator,
  \item survivor selection,
  \item initialization procedure and
  \item termination condition.
\end{itemize}

\noindent In this section these components and features are presented in detail.

In HSA-EA, an individual $\mathbf{y^{(t)}}$ consists of problem variables $y^{(t)}_{1},\ldots,y^{(t)}_{n}$ encoding the values of
weights that determine the initial order in which the vertices are colored, and the control variables $q^{(t)}_{1},\ldots,q^{(t)}_{n}$
denoting the mutation strengths that are used by the operator of normally distributed mutation~\cite{Baeck:1996}. The number of
variables $n$ is equivalent to the number of graph vertices. The problem variables can take the values from the domain $y^{(t)}_{i}\in
Y$, where $Y \in [0.1,1]$. Note that weights cannot reach the value of zero because this value would imply that no constraints exist for the coloring of the corresponding vertex. The control variables can take the values $q^{(t)}_{i}\in[\epsilon_{0},1]$, where the constant $\epsilon_{0}$ is a predefined minimum value of $q^{(t)}_{i}$.

The evaluation of the objective function value is based on Eq.~\ref{eq:1} that counts the number of uncolored vertices. Note that this
function admits an occurrence of neutral solutions as well~\cite{Kimura:1968}. HSA-EA for graph 3-coloring uses the generational
population model $(\mu,\lambda)$, where the whole population is replaced in each generation. Specifically, $\mu$ selected parents
produce $\lambda$ offspring and the ratio $\mu / \lambda = 7$ is used, as suggested in~\cite{Eiben:2003}. An additional individual that
represents the best solution found so far is added to the population of $\mu$ members. This solution is called the reference solution
$\mathbf{y^{*}}$. The tournament selection is used as the parent selection operator and relies on the ordering relation that can rank
any $k$ individuals~\cite{Eiben:2003}. Here, $k$ represents the tournament size.

In self-adaptation, most frequently a normally distributed mutation~\cite{Baeck:1996}, is used that changes the weights as follows:

\begin{equation}
\label{eq:3}
 q^{(t+1)}_{i}=q^{(t)}_{i}\cdot exp(\tau'\cdot N(0,1)+\tau\cdot N_{i}(0,1)),
\end{equation}

\noindent and

\begin{equation}
\label{eq:4}
 y^{(t+1)}_{i}=y^{(t)}_{i}+q^{(t+1)}_{i}\cdot N_{i}(0,1).
\end{equation}

The mutation described in (\ref{eq:3}) and (\ref{eq:4}) is also called uncorrelated mutation with $n$ step sizes~\cite{Eiben:2003}. In
addition to the weights $y^{(t)}_{i}$ of dimension $n$, this kind of mutation requires the mutation strengths $q^{(t)}_{i}$ of the
same dimension. The mutation strengths $q^{(t)}_{i}$ determine an area around the particular weight $y^{(t)}_{i}$ in which the search
process could progress. The parameters $\tau \propto 1/\sqrt{2 \cdot \sqrt{n}}$ and $\tau' \propto 1/ \sqrt{2 \cdot n}$ designate the
learning rate~\cite{Baeck:1996}. The values for mutation strengths can be reduced to zero by the multiplication process in (\ref{eq:4}). In that case, the evolutionary process stagnates. The following condition needs to be considered to prevent this event:

\begin{equation}
\label{eq:5}
 q^{(t+1)}_{i} < \epsilon_{0} \Rightarrow q^{(t+1)}_{i} = \epsilon_{0}.
\end{equation}

Uncorrelated normally distributed mutation with $n$ step sizes can be described as a multiplicative process that decreases mutation
strengths over generations. The mutation strengths determine how big change of weights can be made. The proper setting of the
strengths has a great influence on the exploration of the search space. On the other hand, the parameter value depends on the number of
generations. The higher the number of generations, the bigger the required initial mutation strengths.

The neutral survivor selection operator was developed that represents the hybridization of this evolutionary algorithm and is
presented in Section~\ref{sec:3.4}. The solutions within the population are initialized randomly except for the first solution. For initialization of this solution, a heuristic initialization procedure is used that is explained latter in this section. The algorithm terminates when the maximum number of objective function evaluations is reached or a proper graph 3-coloring is found.

\subsection{Hybrid genotype-phenotype mapping}

Usually, the original problem context in evolutionary algorithm is distinguished from the search space, where the evolutionary process takes place~\cite{Eiben:2003}. Candidate solutions in the original problem space are referred to as phenotypes, while their encodings form the genotypes and represent points in the search space. That is, candidate solutions need to be decoded from their representation, i.e., mapped to the phenotype before they are evaluated. This mapping is also known as a genotype-phenotype mapping. Note that genotype-phenotype mapping is not injective because several genotypes can be mapped into the same phenotype. In line with this, a lot of neutral solutions can appear. Actually, the phenotypes only depend on problem variables, while control parameters determine the manner  in which the genotype space is explored. It can be said that the control parameters describe the strategy for exploring the genotype space and can direct the evolutionary search to new undiscovered regions of the search space.

This genotype-phenotype mapping consists of two steps: transformation of weights to the permutation of vertices and decoding of the solution by the DSatur algorithm. In the former step, from the encoded values of weights, i.e., tuple $\mathbf{y^{(t)}_{i}}= \langle y^{(t)}_{i,1}, \ldots,y^{(t)}_{i,n},q^{(t)}_{i,1}, \ldots ,q^{(t)}_{i,n} \rangle$, $i=1 \ldots \mu$, where $\mu$ denotes the population size, the initial permutation of the vertices is built $\mathbf{v^{(t)}_{i}}=\{v^{(t)}_{i,j}\}$ for $j=1 \ldots n$. The permutation determines the ordering in which the vertices are colored. That is, the vertices are ordered according to the descending values of the corresponding weights. Note that the mutation steps are not used in this transformation. In the latter step, the DSatur algorithm is taken as the construction heuristic by HSA-EA that from the permutation of vertices decodes a coloring $\mathbf{s^{(t)}_{i}}=\{s^{(t)}_{i,j}\}$, where $s^{(t)}_{i,j} \in \{1,2,3\}$. The order of coloring is determined by the DSatur heuristic, as follows:

\begin{enumerate}
 \item The heuristic selects the vertex with the highest saturation, and colors it with the lowest of the three colors.
 \item In the case of a tie, the heuristic selects the vertex with the maximal weight.
 \item In the case of a tie, the heuristic selects a vertex randomly.
\end{enumerate}

The main difference between this heuristic and the original DSatur algorithm is in the second step where the heuristic selects the
vertices according to the weights instead of degrees. Note that the genotype space determined by permutation of the vertices is huge, i.e., $n!$. Therefore, $(n-1)!$ solutions can have equal values of the objective function because only $n$ values of the function exist according to Eq.~\ref{eq:1}. That is, many solutions with the same fitness value can arise that represent neutral networks~\cite{Merz:1999} in the fitness landscape. On the other hand, the phenotype space is much smaller than the genotype space, namely $|C|=3^{n}$.

Initially, the original DSatur algorithm orders the vertices $v_{i,j}\in V$, $j=1 \ldots n$, of a given graph $G$ descending according
to the vertex degrees $deg_{G}(v_{i,j})$ that count the number of edges incident with vertex $v_{i,j}$~\cite{Bondy:2008}. To simulate
the behavior of the original DSatur algorithm~\cite{Brelaz:1979}, the first solution in the population is initialized as follows:
\begin{equation}\label{eq:init}
 y_{i,j}^{(0)}=\frac{ deg_{G}(v_{i,j})}{\textup{max}_{j=1\ldots n}deg_{G}(v_{i,j})}, \ \ \ \ \textup{for}\ j=1 \ldots n.
\end{equation}
Because the genotype is mapped into a permutation of weights, the same ordering is obtained as by the original DSatur, where the
solution can be found in one step.

\subsection{Local search heuristic}

Experiments with self-adaptive evolutionary algorithms prove that a search can rapidly converge towards good areas of the search space
(exploration)~\cite{Stadler:1995}. During the exploitation phase, these algorithms are less convenient because of the stochastic nature of the variation operators. In this phase, improvement heuristics are useful that may improve current solutions with a more systematic
search in their vicinity. Local search~\cite{Hoos:2005} is the most often used kind of improvement heuristic that can incorporate
problem-specific knowledge into the evolutionary algorithm.

Local search can be described as an iterative heuristic that explores a set of candidate solutions around the current solution (also
known as neighborhood) and can replace the current solution with a better one, if it is found. A neighborhood of the current solution
$\mathbf{y}$ is a mapping $\mathcal{N}:Y \mapsto 2^{Y}$, which for each solution $\mathbf{y} \in Y$ defines a set
$\mathcal{N}(\mathbf{y}) \subseteq Y$ of solutions that can be reached using a unary operator~\cite{Aarts:1997}. In fact, each solution
in the neighborhood $\mathbf{y'} \in \mathcal{N}(\mathbf{y})$ can be reached from the current solution $\mathbf{y}$ in $k$ steps. Therefore, this neighborhood is also called the $k$-opt neighborhood of the current solution $\mathbf{y}$.

\begin{figure}[htb!]
\centering
\includegraphics [scale=0.39]{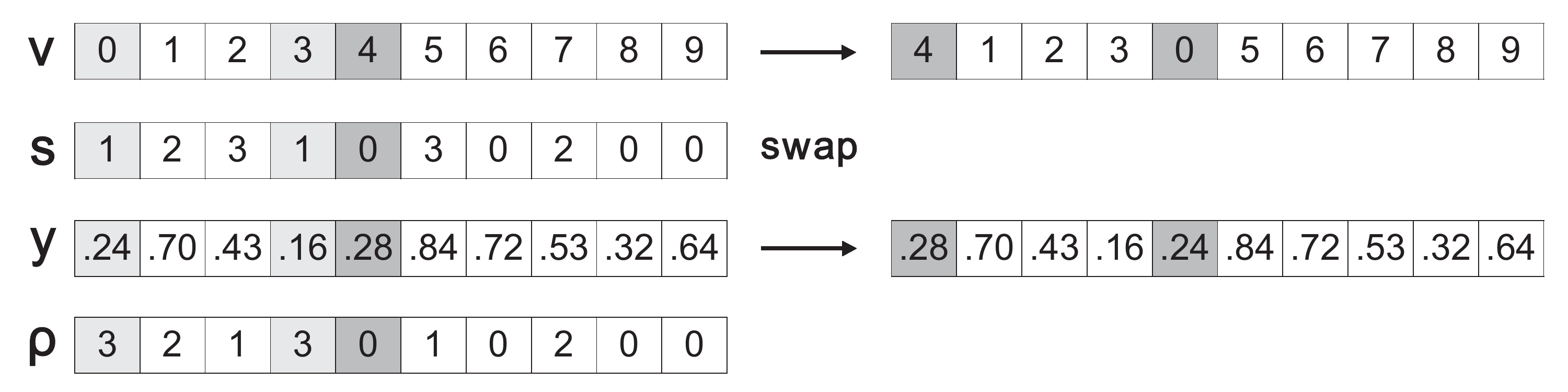}
\caption{The hybrid swap unary operator}
\label{fig:Swap}
\end{figure}

Although various unary operators have been developed to be used with HSA-EA, experiments have shown that the best performance can be achieved by the unary operator termed the \textit{hybrid swap}. Therefore, this operator was used in this work. The functioning of this operator is illustrated in Fig.~\ref{fig:Swap}, which presents a solution to a graph $G$ with ten vertices. This solution is composed of a permutation of vertices $\mathbf{v}$, corresponding coloring $\mathbf{s}$, weights $\mathbf{y}$, and saturation degrees $\rho$. The hybrid swap unary operator takes the first uncolored vertex in a solution and orders the predecessors according to the saturation degree, in descending order. The uncolored vertex is swapped with the vertex that has the highest saturation degree. In the case of a tie, the operator randomly selects a vertex among the vertices with higher saturation degrees. Thus, the best neighbor of the current solution is determined by an exchange of two vertices ($1$-opt neighborhood).

In Fig.~\ref{fig:Swap}, an element of the solution corresponding to the first uncolored vertex 4 is in dark grey and the vertices 0 and 3 with the highest saturation degree are in light grey. From vertices 0 and 3, hybrid swap randomly selects vertex 0 and swaps it with vertex 4 (the right-hand side of Fig.~\ref{fig:Swap}).

\subsection{Neutral survivor selection}
\label{sec:3.4}

A genotype diversity is one of the main prerequisites for efficient self-adaptation. The smaller genotypic diversity causes the
population to be crowded within the search space. As a result, the search space is exploited. On the other hand, the larger genotypic diversity causes the population to be better distributed over the search space and therefore, the search space is better
explored~\cite{Baeck:1996}. The genotype diversity is explicitly maintained using the proposed neutral survivor selection that is
inspired by the neutral theory of molecular evolution~\cite{Kimura:1968}.

However, a measure is needed to determine the distance between solutions. An Euclidian distance is the most appropriate when the solutions are represented as real-coded vectors. The Euclidian distance between two vectors $\mathbf{y_{1}}$ and $\mathbf{y_{2}}$ is
expressed as follows:
\begin{equation}
\label{eq:Euclid}
 d_{E}(\mathbf{y_{1}},\mathbf{y_{2}})=\sqrt{\frac{1}{n}\cdot \sum_{j=1}^{n}(y_{1,j}-y_{2,j})^{2}},
\end{equation}
A reference solution is needed to determine how the solutions are dissipated over the search space. For this reason, the current best
solution $\mathbf{y^{*}}$ in the population is used, as defined in Subsection~\ref{ssec:3}.

The operation of the neutral survivor selection is divided into two phases. During the first phase, in the population of $\lambda$ offspring the evolutionary algorithm finds the set of the best neutral solutions $N_{S}=\{\mathbf{y_{1}}, \ldots ,\mathbf{y_{k}}\}$. If the neutral solutions are better than, or equal, to the reference solution, i.e., $f(\mathbf{y_{i}}) \leq f(\mathbf{y^{*}})$ for $i=1, \ldots ,k$, the reference solution $\mathbf{y^{*}}$ is replaced by the neutral solution $\mathbf{y_{i}} \in N_{S}$ that is most distant from the reference solution according to Eq.~\ref{eq:Euclid}. In line with this, it is expected that the evolutionary search is directed towards new, undiscovered regions of the search space. Conversely, if the neutral solutions are worse than the reference solution, the reference solution remains unchanged.

During the second phase, the reference solution $\mathbf{y^{*}}$ is used for determining the next population of survivors. For this purpose, the offspring are ordered according to the ordering relation $\prec$ (read: is better than) as follows:
\begin{equation}
\label{eq:order}
 f(\mathbf{y_{1}}) \prec \ldots \prec f(\mathbf{y_{i}}) \prec f(\mathbf{y_{i+1}}) \prec \ldots \prec f(\mathbf{y_{\lambda}}),
\end{equation}

\noindent where the ordering relation $\prec$ is defined as

\begin{equation}
\label{eq:order2}
f(\mathbf{y_{i}})\prec f(\mathbf{y_{i+1}})\Rightarrow \left\{\begin{matrix}
f(\mathbf{y_{i}})< f(\mathbf{y_{i+1}}), \\
f(\mathbf{y_{i}})= f(\mathbf{y_{i+1}})\wedge (d(\mathbf{y_{i}},\mathbf{y^{*}})>d(\mathbf{y_{i+1}},\mathbf{y^{*}})).
\end{matrix}\right.
\end{equation}

Finally, for the next generation the evolutionary algorithm selects the best $\mu$ offspring according to Eq.~\ref{eq:order}. These
individuals take random positions in the next generation. Like the neutral theory of molecular evolution, the neutral survivor selection offers three possible outcomes to offspring, as follows. First, the best offspring survive. Additionally, the neutral solution that is the most distant from the reference solution becomes the new reference solution. Second, the low-fitness offspring are usually eliminated from the population. Third, all other solutions, that could be neutral as well, can survive if they take the first $\mu$ positions according to Eq.~\ref{eq:order}.

\section{Experiments and results}

The goal of the experiments performed in this study was to compare the results of different graph 3-coloring algorithms and show that
the performance of the proposed HSA-EA for graph 3-coloring is comparable with, or better than, the performance of other algorithms for solving this problem. Here, the following algorithms were used:
\begin{itemize}
  \item SAW-EA~\cite{website:Hemert},
  \item Tabucol~\cite{website:Chiarandini},
  \item HEA~\cite{website:Chiarandini},
  \item HSA-EA.
\end{itemize}

An implementation of SAW-EA can be found in~\cite{website:Hemert} and is described in~\cite{Eiben:1998}. The Tabucol and HEA
algorithms are among the best known algorithms for solving $k$-GCP~\cite{deWerra:1985,Galinier:1999}. In addition, implementations of both algorithms are publicly available for facilitating comparisons between newly developed algorithms~\cite{Chiarandini:2010}. Although the implementation of Tabucol, as presented in~\cite{Hamiez:2010}, gained slightly better results than in~\cite{website:Chiarandini}, the former implementation is not publicly available and, therefore, was not used in our study.

The characteristics of HSA-EA used in the experiments were as follows. The fitness function according to Eq.~\ref{eq:1} was considered by this algorithm. Note that the fitness value cannot exceed the number of vertices since the fitness function counts the number of vertices violating the constraints. This value, however, cannot be higher than the number of all vertices. HSA-EA employed the selection scheme $(15,100)$. That is, 100 candidate solutions were generated from the population of 15 parents, from which 15 fittest offspring were selected by the neutral survivor selection. Although other selection schemes were also examined, the experiments showed that this selection scheme was the most effective. Typically, a larger population requires higher initial mutation strengths. Furthermore, a larger population can maintain higher diversity, but a smaller population converges faster than the larger one. In our opinion, this is due to the fact that the used predetermined number of evaluations is insufficient for the larger population to converge. Experiments varying the tournament size showed that on average the best results are obtained using a tournament size of 3. However, a detailed analysis of the experimental results proved that a tournament size of 3 is more appropriate for smaller populations (such as in our case). As the population size grows, choosing a tournament size of 5 is more suitable. Determination of
initial mutation strengths was crucial for the performance of HSA-EA. After extensive experimentation, the initial mutation strength $s^{(0)}_{i}=0.03$ and $\epsilon_{0}=0.001$ produced the best results. As the termination condition, the number of function
evaluations was limited to 300,000, while the number of independent runs was fixed at 25. The last two settings were the same as
in~\cite{Eiben:1998}.

The algorithms were compared according to two performance measures:
\begin{itemize}
  \item success rate (SR) and
  \item average number of objective function evaluations to solution (AES).
\end{itemize}

The former reflects the stochastic nature of HSA-EA and is defined as the ratio of successfully terminated runs, while the latter expresses the efficiency of a particular algorithm, and is defined as the number of objective function evaluations needed to find a solution.

\subsection{Test problems}

Heuristics for solving the $k$-GCP are commonly compared on a set of graphs as proposed in the DIMACS challenge
suite~\cite{Johnson:1996}. However, this set does not permit the statistical study of relations between algorithm performance and
the structural features of graphs~\cite{Chiarandini:2010}. The structural features of graphs may be induced by:
\begin{itemize}
  \item generating a graph with a given number of color classes and edge densities,
  \item influencing the variability in the sizes of the color classes.
\end{itemize}
Both presumptions can be satisfied using the Culberson graph generator~\cite{website:Culberson}. Although this graph generator
can generate various types of \textit{k}-colorable graphs, we focus on three specific types: uniform, equi-partite, and flat graphs. The type of graphs imposes an edge distribution by the graph generator as follows. Uniform 3-colorable graphs are random graphs, where the vertices are randomly assigned to one of the three color classes uniformly and independently. The main property of equi-partite 3-colorable graphs is that the three color classes are as equal in size as possible. This type of graphs is less difficult to 3-color but difficult enough for many existing algorithms~\cite{Culberson:1996}. Flat graphs seem the hardest to 3-color because besides the minimum variation in number of vertices in the three color classes the variation in degree for each node is also kept to minimum.

Random graphs are created using the parameter $\Delta \in \{0, 1, 2\}$ that controls the variability in sizes of the color classes.
Uniform $3$-colorable graphs are generated when $\Delta = 0$ as mentioned previously. Variable $3$-colorable graphs are obtained when $\Delta > 0$. Graphs of this type are generated as follows. For each vertex in turn, the generator randomly selects an integer $r$ from the interval $[0,\Delta]$. Then, this vertex is assigned to the color class $i$ that is randomly selected from the interval $[r,2]$. In general, variable $3$-colorable graphs are easier to color, as reported by Turner~\cite{Turner:1988}.

The following parameters were used when generating the graphs: graph type, number of vertices, edge density controlled by the parameter $p$ determining the probability that two vertices $v_{i}$ and $v_{j}$ are connected with an edge $(v_{i},v_{j})$, and the seed of the random graph generator. In general, these graphs can be denoted as $G_{t,n,p,s}$, where \textit{t} denotes the graph type (i.e.,
\textit{uni} for uniform ($\Delta = 0$), \textit{eq} for equi-partite, \textit{flat} for flat), $n$ the number of vertices, $p$ the
probability controlling the edge density, and $s$ the seed of the random graph generator. Three types of uniform graphs were generated in order to investigate the influence of variability in sizes of the color classes on the algorithm results. When $\Delta =0$, the
generated graphs have no variability in the sizes of the color classes (the classes tend to be nearly equal in size, like in the equi-partite graphs), while when $\Delta=1$ or $\Delta=2$, the sizes tend to vary considerably. Flat graphs can be generated using different flatness when determining variations in the degrees of the vertices. If the flatness is zero, the generated graphs are fairly uniform and these graphs are the hardest to color. This type of flat graph was used in our experiments.

Graphs with the number of vertices $n=500$ and $n=1,000$, were considered during the experiments. These graphs are denoted as medium-scale and large-scale graphs in this paper. Note that the graphs with the number of vertices $n=200$ as also treated in~\cite{Eiben:1998} were not considered in this study because these graphs are too simple for the tested graph 3-coloring algorithms. The random graph generator seed parameter was varied from 1 to 10 with a step of one in order to test the statistical significance of the results. Additionally, the seeds of the random number generator for all the used algorithms were different in each run.

Most combinatorial optimization problems are sensitive to the phase transition which refers to the regions where the problem passes from the state of ``solvable'' to the state ``unsolvable'', and vice versa~\cite{Turner:1988}. Typically, these regions are determined by a problem-specific parameter. For 3-GCP this parameter is the edge density determined by probability that two vertices are connected, $p$. Many authors have identified various critical values for this parameter when determining the phase transition region. For example, Petford and Welsh~\cite{Petford:1989} stated that this phenomenon occurs when $2pn/3 \approx 16/3$, Cheeseman et al.~\cite{Cheeseman:1991} $2m/n \approx 5.4$, Eiben et al.~\cite{Eiben:1998} $7/n \leq p \leq 8/n$, and Hayes~\cite{Hayes:2003} $m/n \approx 2.35$. The parameter $m$ in the above formulas denotes the number of edges.

In order to capture the phase transition, the parameter $p$ needs to be varied appropriately. Preliminary experiments were conducted to determine suitable values of the value $p$. The following experimental setup was applied to show that the phase transition was located correctly. Medium- and large-scale graphs were generated with edge density $p \in [0.004, 0.696]$ with a step of 0.004. As a result, 174 instances were obtained for each type of graphs. Specifically, random graphs were generated with different variabilities in size, i.e., with $\Delta \in \{ 0, 1, 2 \}$. Graphs of each type were generated varying random graph generator seed from 1 to 10. Then, the four algorithms, i.e., SAW-EA, Tabucol, HEA, and HSA-EA, run the graph 3-coloring and the results on random graphs with different seeds were accumulated over 25 runs. The results of this experiment showed that none of the tested algorithms had problems with the 3-coloring of medium-scale graphs with $p>0.028$ and large-scale graphs with $p>0.014$. For this reason, our further experiments were conducted with these two values as upper bounds of the observed edge densities.

Essentially, the experiments were divided into three parts. In the first part, the behavior of the mentioned graph-coloring algorithms was observed in the phase transition region. This part is comparable to the study of Eiben et al. in~\cite{Eiben:1998}. The influence of the evolutionary algorithm on the traditional DSatur heuristic was observed during the second part. The structural features of graphs were studied during the third part. This part considered the novel experimental approach to graph coloring, as proposed by Chiarandini and St\"{u}tzle~\cite{Chiarandini:2010}.

\subsection{Behavior of graph-coloring algorithms in the phase transition}

These experiments were executed on medium- and large-scale graphs. The results for both graph sizes are presented in the next sections. In both cases, however, the phenomenon of the phase transition was captured.

\subsubsection{Medium-scale graphs}
\label{sub_sec:medium}

Medium-scale graphs ($n=500$) were generated with edge densities varied from $p=0.008$ to $p=0.028$ with a step of $0.001$. As a result, 21 instances of randomly created graphs were obtained. Note that the phase transition occurs at $p=0.014$ according to Hayes~\cite{Hayes:2003}, $p=0.016$ according to Cheeseman~\cite{Cheeseman:1991}, and Petford and Welsh~\cite{Petford:1989}, and $p \in [0.014,0.016]$ according to Eiben et al.~\cite{Eiben:1998}.

\begin{SCfigure}
\centering
\label{fig:Fig_UMS}
\caption{SR on uniform medium-scale graphs}
\includegraphics[width=0.7\textwidth]{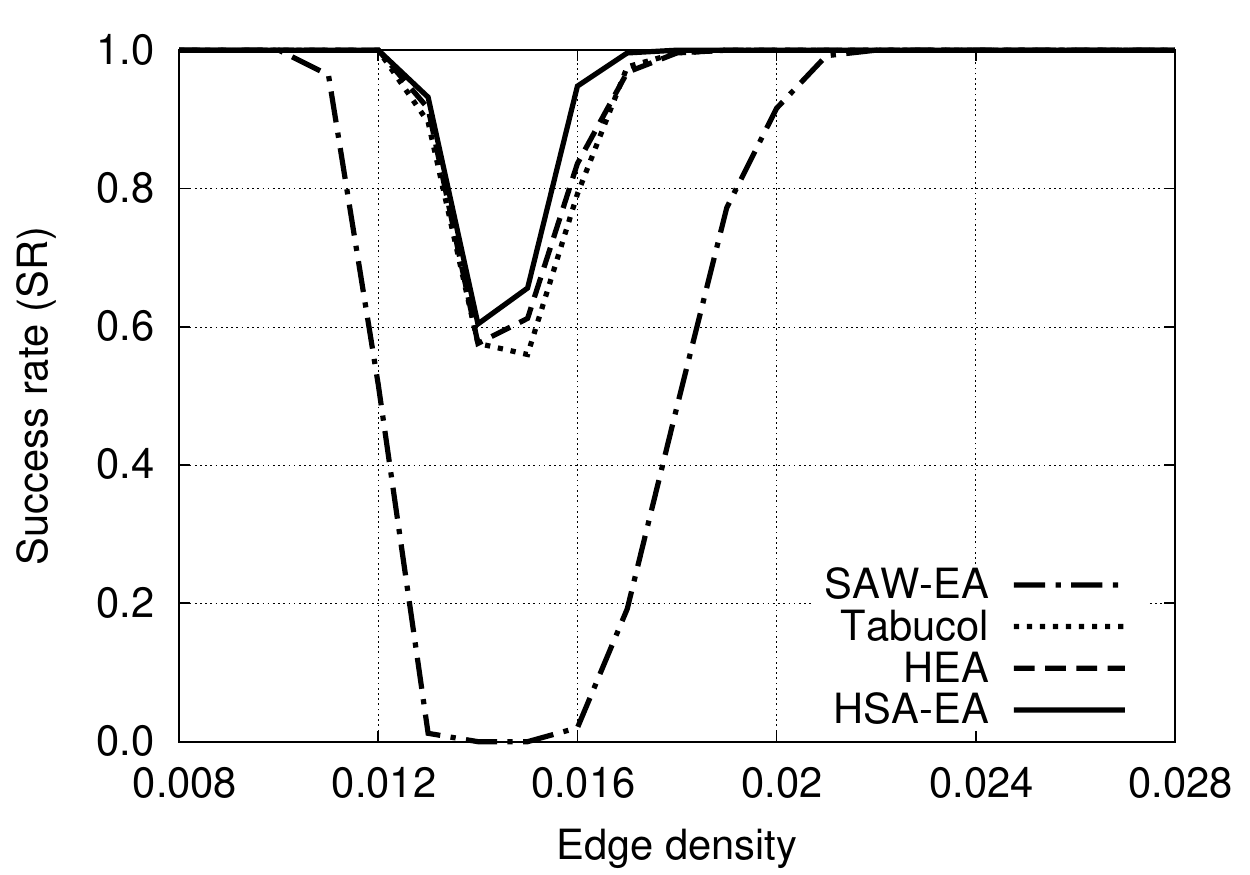}
\vspace{-5mm}
\end{SCfigure}

\begin{SCfigure}
\centering
\label{fig:Fig_UMA}
\caption{AES on uniform medium-scale graphs}
\includegraphics[width=0.7\textwidth]{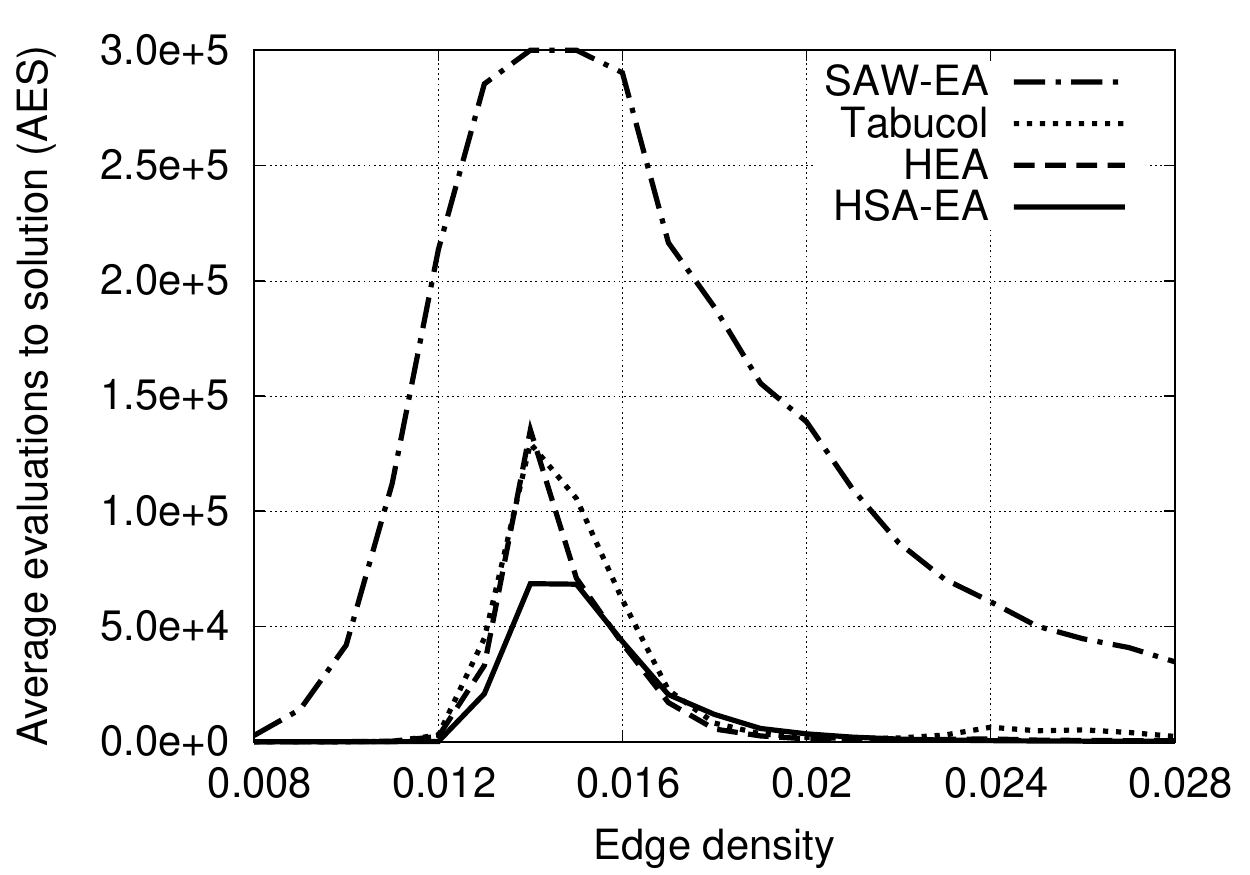}
\vspace{-5mm}
\end{SCfigure}

\begin{SCfigure}
\centering
\label{fig:Fig_EMS}
\caption{SR on equi-partite medium-scale graphs}
\includegraphics[width=0.7\textwidth]{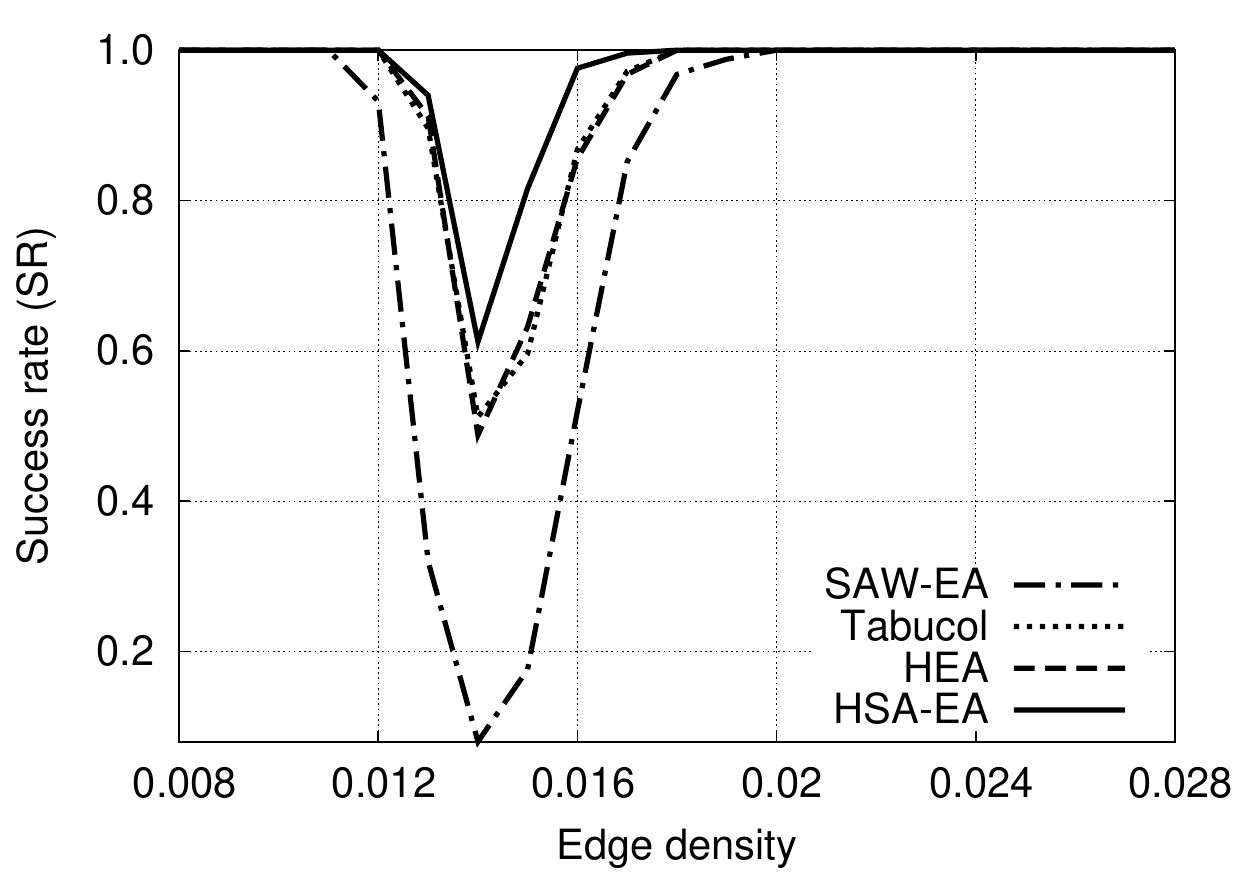}
\vspace{-5mm}
\end{SCfigure}

\begin{SCfigure}
\centering
\label{fig:Fig_EMA}
\caption{AES on equi-partite medium-scale graphs}
\includegraphics[width=0.7\textwidth]{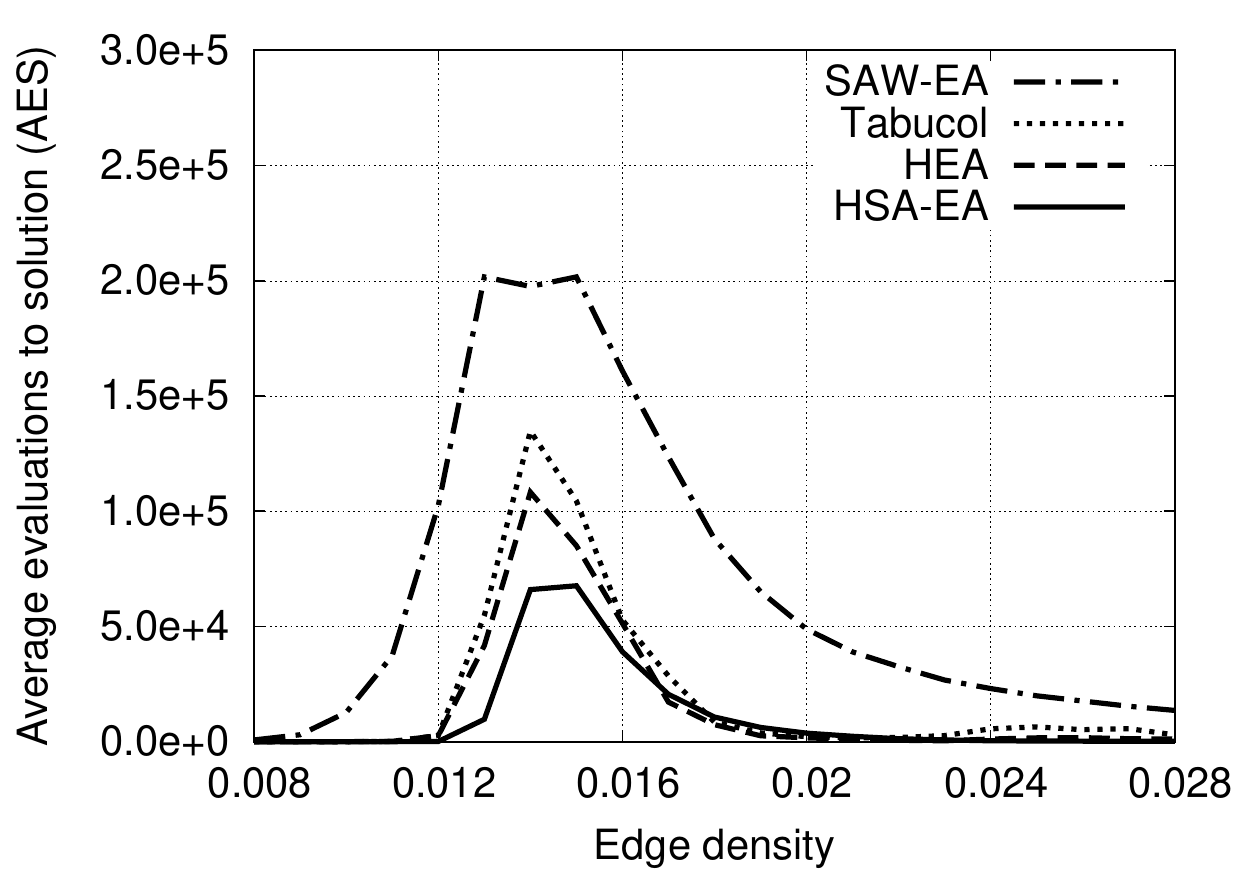}
\vspace{-5mm}
\end{SCfigure}

\begin{SCfigure}
\centering
\label{fig:Fig_FMS}
\caption{SR on flat medium-scale graphs}
\includegraphics[width=0.7\textwidth]{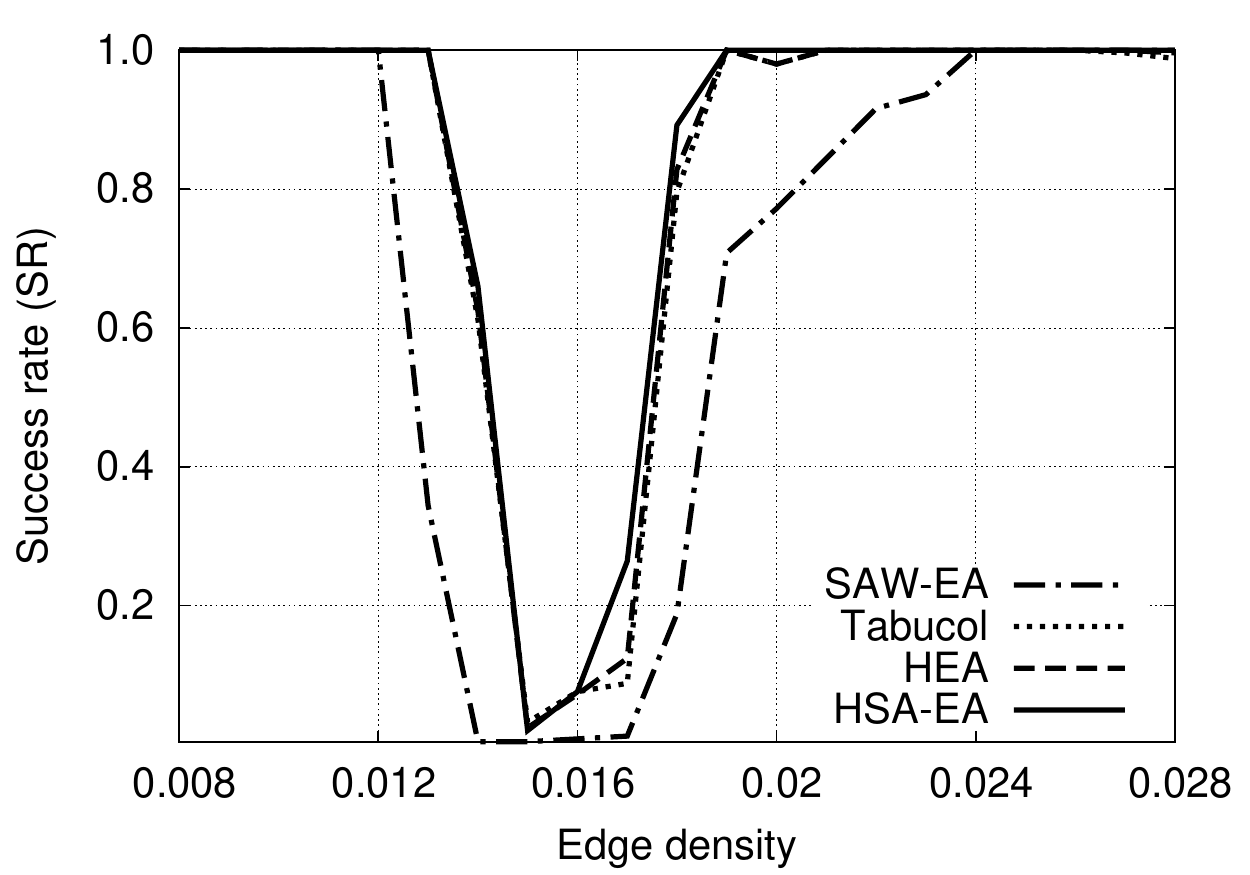}
\vspace{-5mm}
\end{SCfigure}

\begin{SCfigure}
\centering
\label{fig:Fig_FMA}
\caption{AES on flat medium-scale graphs}
\includegraphics[width=0.7\textwidth]{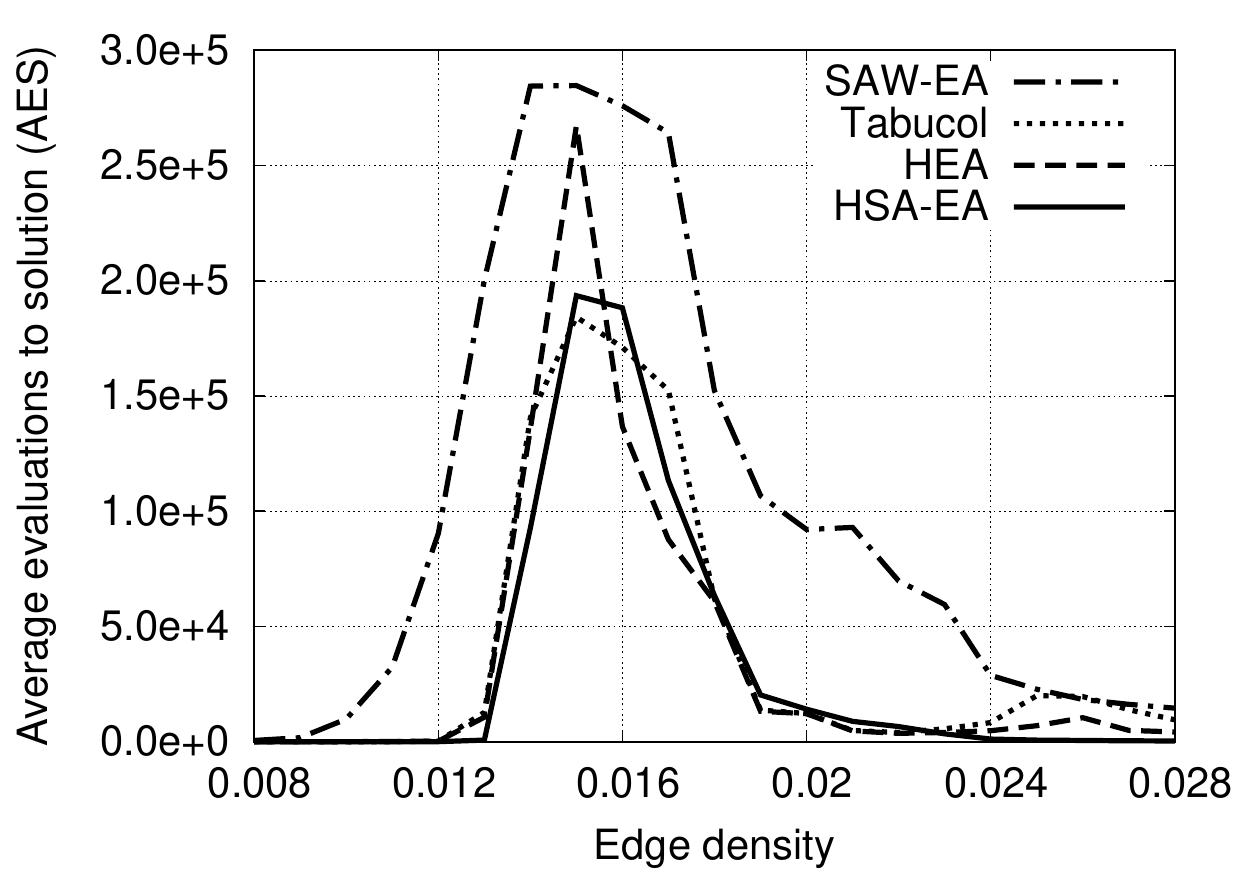}
\vspace{-5mm}
\end{SCfigure}

The averaged results of the tested algorithms on medium-scale graphs with different seeds are illustrated in Figs.~\ref{fig:Fig_UMS}-\ref{fig:Fig_FMA}. These figures represent six diagrams for three graph types according to two measures. The best results on medium-scale graphs were obtained by HSA-EA (Fig.~\ref{fig:Fig_UMS}). The average results of Tabucol and HEA were similar, while SAW-EA gained the worst results according to SR. In fact, this algorithm was sensitive to the phase transition as proposed by Hayes ($p=0.014$). Results on uniform graphs according to AES (Fig.~\ref{fig:Fig_UMA}) confirmed that graphs of this type can be colored by all tested algorithms, except SAW-EA. For this reason, the AES plots did not reach the maximum number of evaluations to solution.

As illustrated in Fig.~\ref{fig:Fig_EMS}, the equi-partite graphs were also best 3-colored by HSA-EA. Note that all the mentioned
algorithms were sensitive to the phase transition determined by $p=0.014$. In this case, all algorithms 3-colored the
equi-partite graphs with $\textnormal{SR}>0$, as seen in the AES plots in Fig.~\ref{fig:Fig_EMA}. Note that SAW-EA achieved
$\textnormal{SR}=0.08$ at $p=0.014$.

Experiments on flat graphs confirmed that these graphs were the hardest to color. Interestingly, the SR plots for these graphs showed single valleys (Fig.~\ref{fig:Fig_FMS}) and the AES plots single peaks (Fig.~\ref{fig:Fig_FMA}). Note that
for SAW-EA, SR did not reach zero on average, i.e., $\textnormal{SR}=0.004$ at $p=0.014$, as can be seen in the corresponding AES
plot. As a matter of fact, the peaks (valleys) were located as suggested by Hayes~\cite{Hayes:2003},
Cheeseman~\cite{Cheeseman:1991}, Petford and Welsh~\cite{Petford:1989}, and Eiben et al.~\cite{Eiben:1998}, i.e., in the interval $p
\in [0.14,0.16]$.

\subsubsection{Large-scale graphs}
\label{sub_sec:large}

Large-scale graphs ($n=1,000$) were generated with edge densities varying from $p=0.004$ to $p=0.014$ with a step of $0.0005$. As a result, 21 instances of randomly generated graphs were obtained for each graph type. The phase transition occurs at $p=0.007$ according to Hayes~\cite{Hayes:2003}, $p=0.008$ according to Cheeseman~\cite{Cheeseman:1991}, and Petford and Welsh~\cite{Petford:1989}, and $p \in [0.007,0.008]$ according to Eiben et al.~\cite{Eiben:1998}.

\begin{SCfigure}
\centering
\label{fig:Fig_ULS}
\caption{SR on uniform large-scale graphs}
\includegraphics[width=0.7\textwidth]{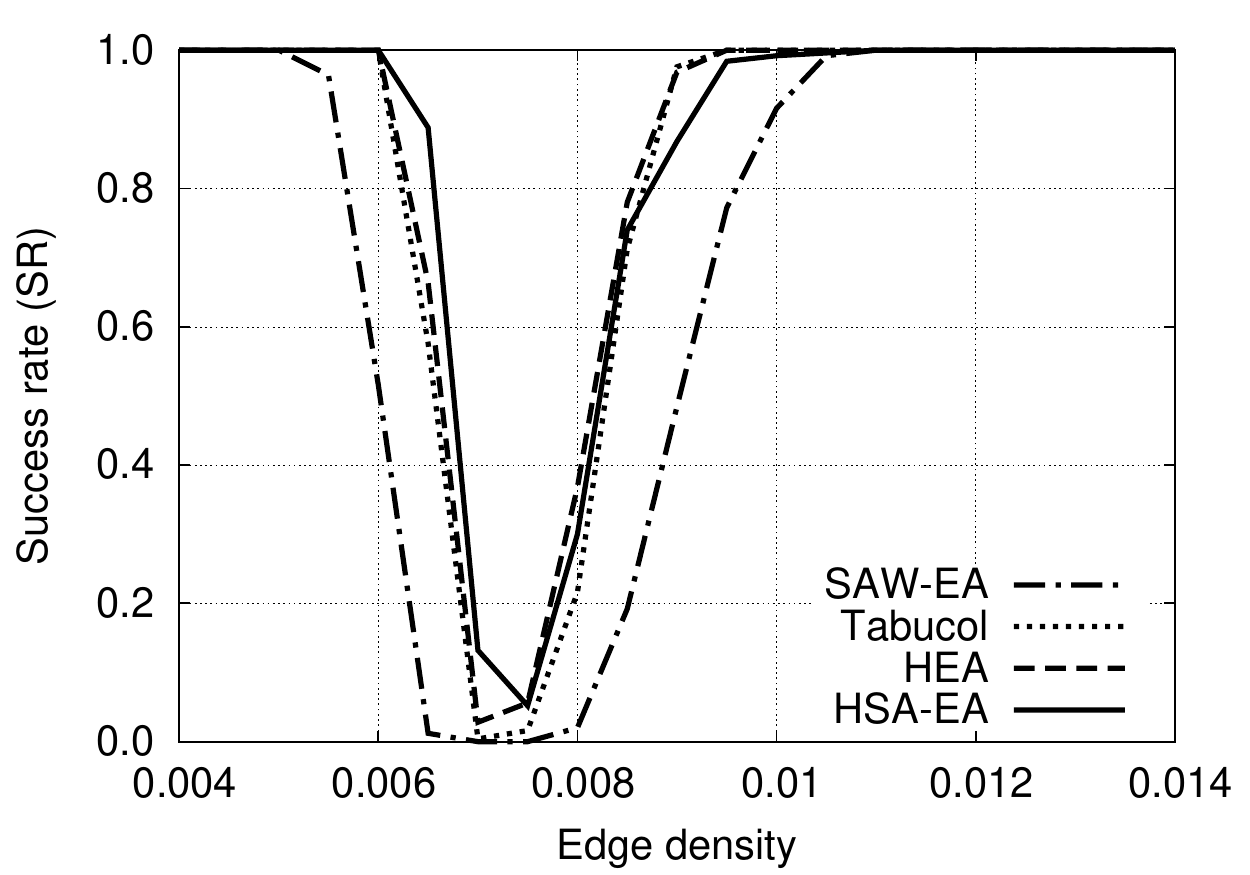}
\vspace{-5mm}
\end{SCfigure}

\begin{SCfigure}
\centering
\label{fig:Fig_ULA}
\caption{AES on uniform large-scale graphs}
\includegraphics[width=0.7\textwidth]{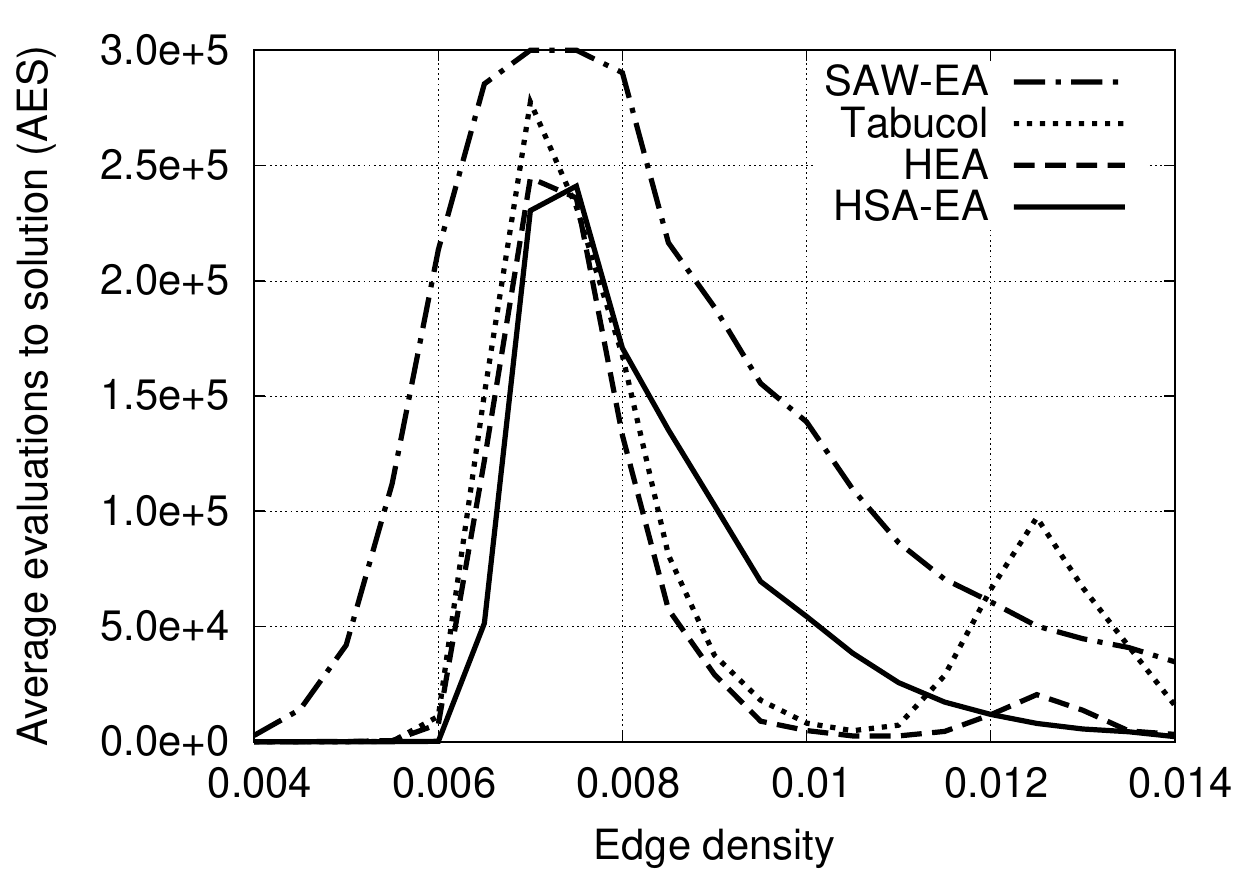}
\vspace{-5mm}
\end{SCfigure}

\begin{SCfigure}
\centering
\label{fig:Fig_ELS}
\caption{SR on equi-partite large-scale graphs}
\includegraphics[width=0.7\textwidth]{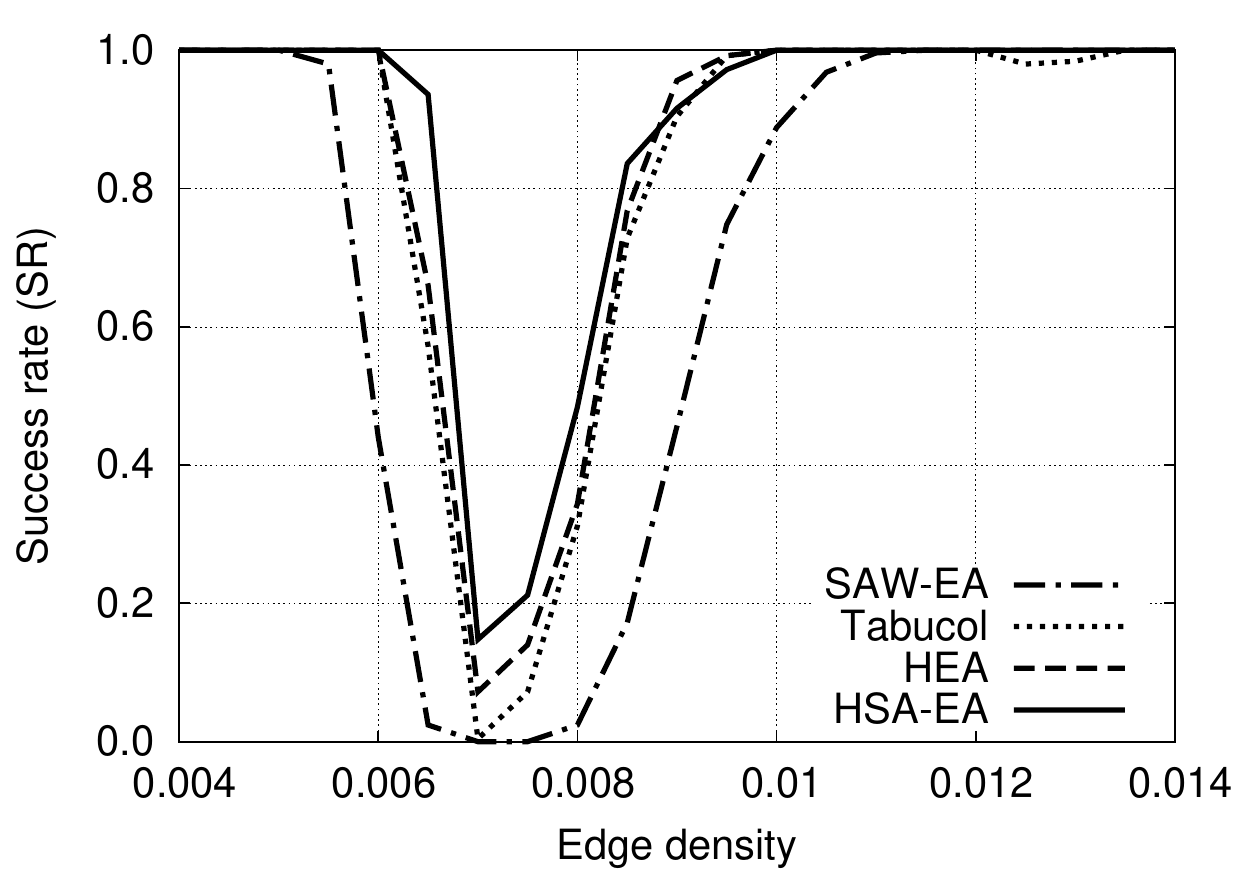}
\vspace{-5mm}
\end{SCfigure}

\begin{SCfigure}
\centering
\label{fig:Fig_ELA}
\caption{AES on equi-partite large-scale graphs}
\includegraphics[width=0.7\textwidth]{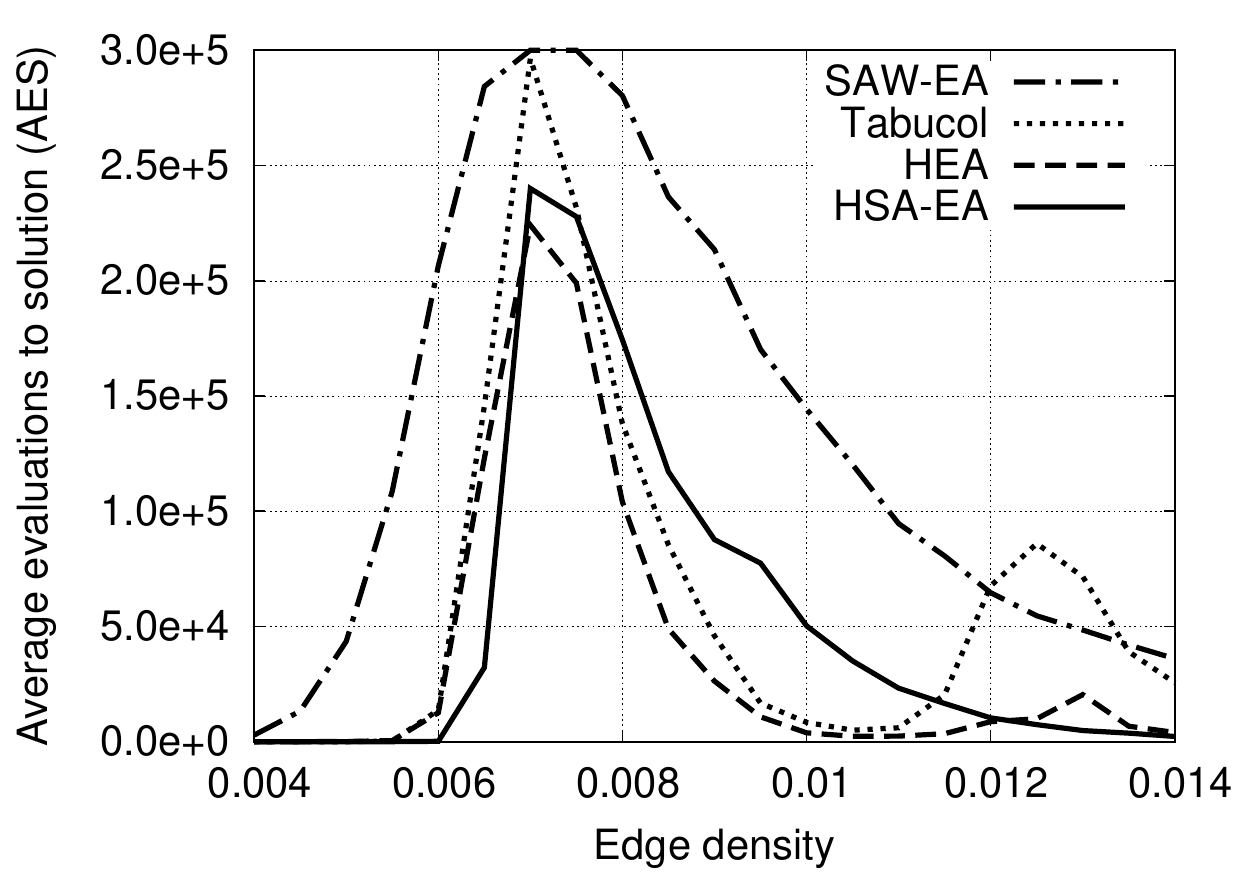}
\vspace{-5mm}
\end{SCfigure}

\begin{SCfigure}
\centering
\label{fig:Fig_FLS}
\caption{SR on flat large-scale graphs}
\includegraphics[width=0.7\textwidth]{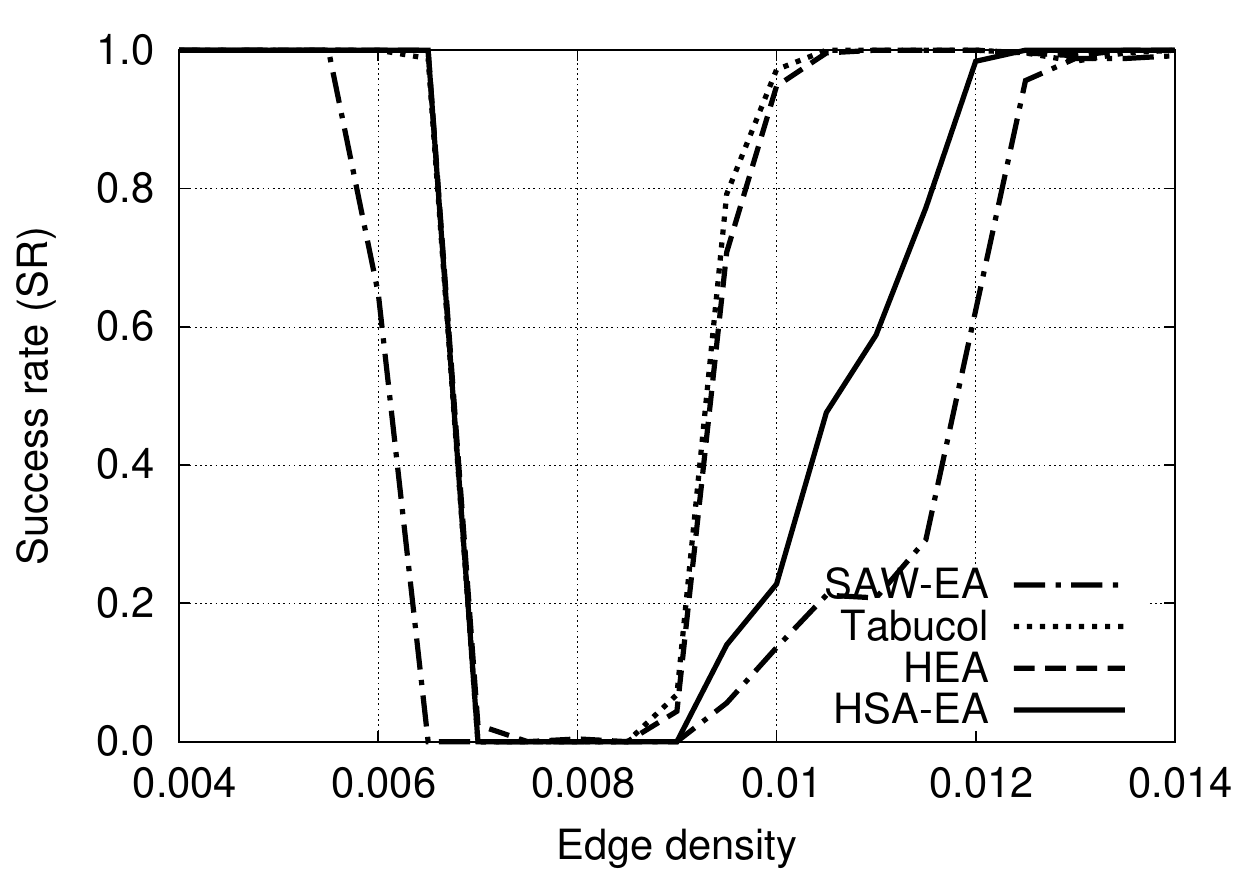}
\vspace{-5mm}
\end{SCfigure}

\begin{SCfigure}
\centering
\label{fig:Fig_FLA}
\caption{AES on flat large-scale graphs}
\includegraphics[width=0.7\textwidth]{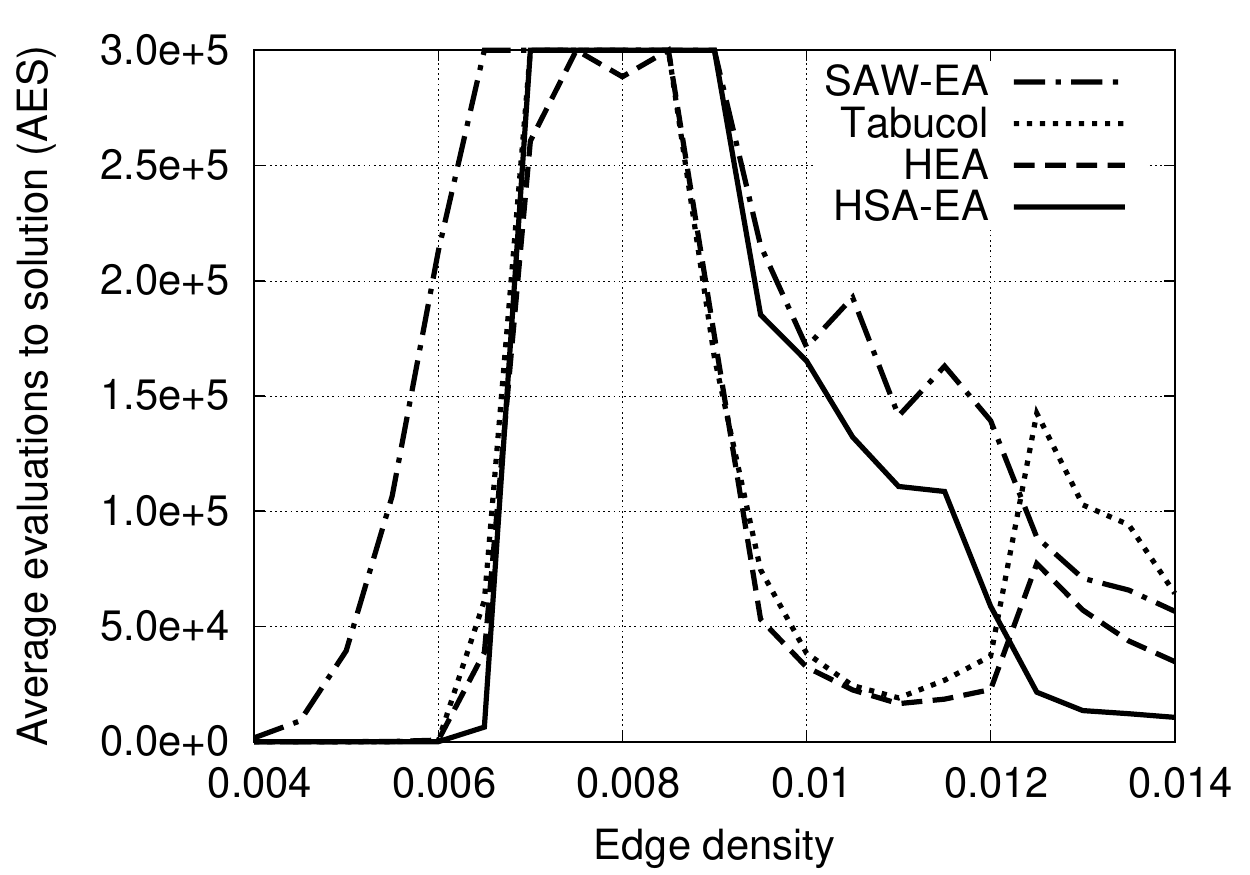}
\vspace{-5mm}
\end{SCfigure}

The results of graph 3-coloring using the tested algorithms are illustrated in Figs.~\ref{fig:Fig_ULS}-\ref{fig:Fig_FLA}. These figures show the results of coloring three graph types according to two measures.

HSA-EA produced the best results for $p<0.007$ when coloring the uniform graphs (Fig.~\ref{fig:Fig_ULS}). On average, graph instances with $p=0.007$ were a hard nut to crack for all the tackled algorithms. However, HSA-EA obtained $\textnormal{SR}=0.16$,
while the other algorithms performed slightly worse. Note that SAW-EA did not find any solution. HEA improved its own results at
$p=0.0075$, and, thus, reached the results of HSA-EA. When the edge density was increased, the best results were
obtained by HEA. Tabucol and HSA-EA produced similar results, while SAW-EA obtained the worst results. According to AES (Fig.~\ref{fig:Fig_ULA}), the best results were produced by HEA that spent the minimum number of evaluations on the
graph instances with edge density away from the phase transition. Similar results were gained by Tabucol except for the graph instance with $p=0.013$, where the AES measure increased significantly.

Equi-partite graphs were most successfully colored by HSA-EA (Fig.~\ref{fig:Fig_ELS}). In general, the SR plot
describing the behavior of a particular algorithm in the phase transition was the narrowest for HSA-EA although HEA obtains
slightly better results according to the AES measure (Fig.~\ref{fig:Fig_ELA}). However, the HEA's and HSA-EA's plots did not reach the
maximum number of evaluations during the observed interval of the edge density. Note that Tabucol did not completely solve all of the equi-partite graph instances in the vicinity of $p=0.013$. This behavior of Tabucol seemed to be related to the occurrence of the second phase transition~\cite{Boettcher:2004}.

The results on flat graphs (Fig.~\ref{fig:Fig_FLS}) indicate that graphs of this type are the hardest to color. Interestingly, the
best coloring algorithms, like HEA and Tabucol, also did not color graph instances in the phase transition. Moreover, the region where these algorithms do not find any solution was widened to $p \in [0.007,0.0085]$. This region was broader for HSA-EA and even broader for SAW-EA. Note that Tabucol and HEA were also sensitive to the second phase transition. According to AES (Fig.~\ref{fig:Fig_FLA}), the worst results in the phase transition region were obtained by SAW-EA. Slightly better results were produced by HSA-EA which, on the other hand, outperformed the other algorithms in the second phase transition region.

\subsection{Comparing HSA-EA with the DSatur traditional algorithm}

To see the contribution on HSA-EA with regard to the original DSatur algorithm, the two algorithms were compared. Note that HSA-EA affects the selection of the first vertex on DSatur heuristic. In order to show how important this decision is, a modified variant of the DSatur algorithm~\cite{Culberson:1996} was built, where only one run is performed similarly to the fitness calculation phase in HSA-EA. As for the first vertex, each of the $n$ vertices in the permutation is selected sequentially. As a result, $n$ different runs of this algorithm (denoted as ModDSat) were performed. Additionally, the results were also compared with the backtracking variant of DSatur~\cite{website:Trick} (denoted as BkDSat). Thanks to tie breaking, this algorithm is also stochastic because different random number generator seed values during the initialization of the algorithm may lead to different results.

Like in Section~4.2, the experiments were conducted on medium-scale and large-scale graphs. Furthermore, the phase transition was also
included.

\subsubsection{Medium-scale graphs}

\begin{SCfigure}
\centering
\label{fig:Fig_MED1}
\caption{SR on uniform medium-scale graphs}
\includegraphics[width=0.7\textwidth]{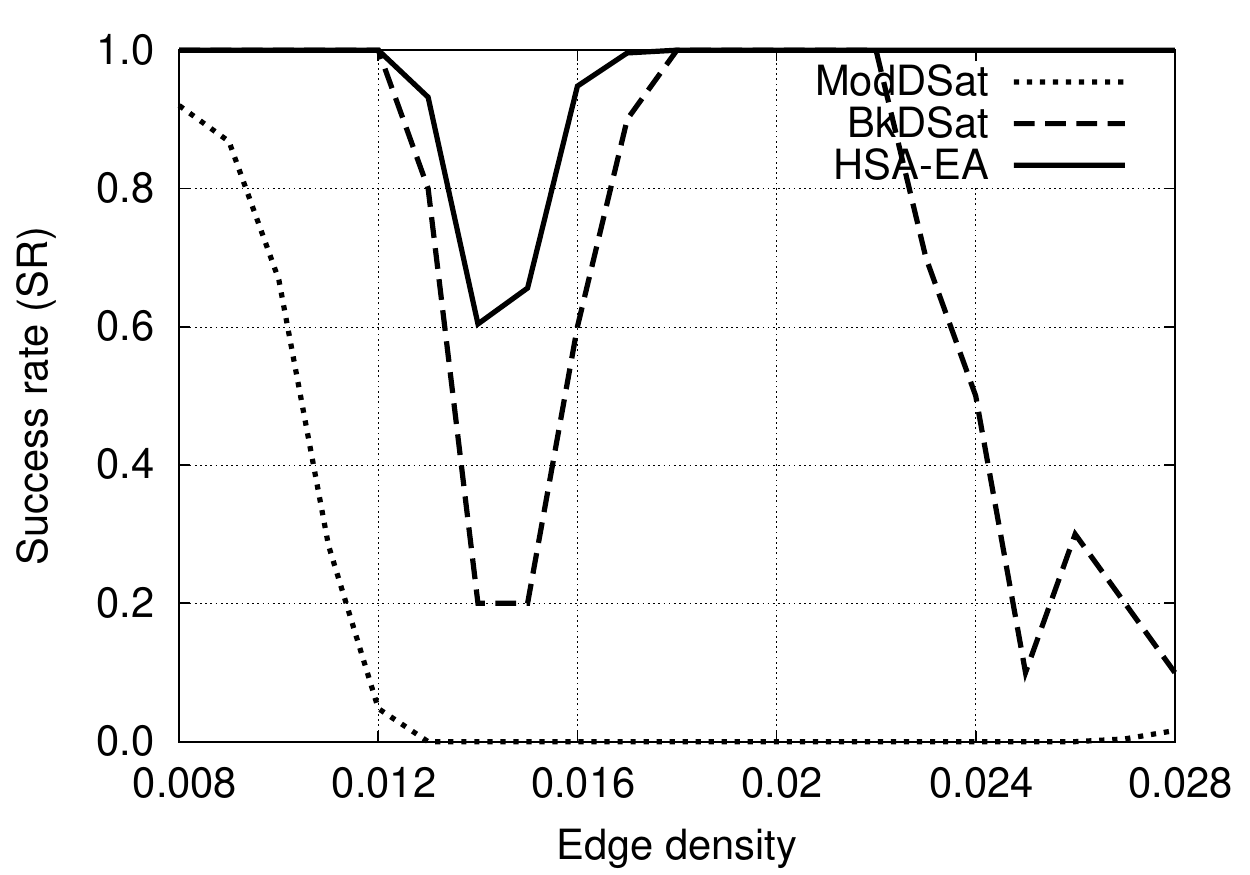}
\vspace{-5mm}
\end{SCfigure}

\begin{SCfigure}
\centering
\label{fig:Fig_MED2}
\caption{SR on equi-partite medium-scale graphs}
\includegraphics[width=0.7\textwidth]{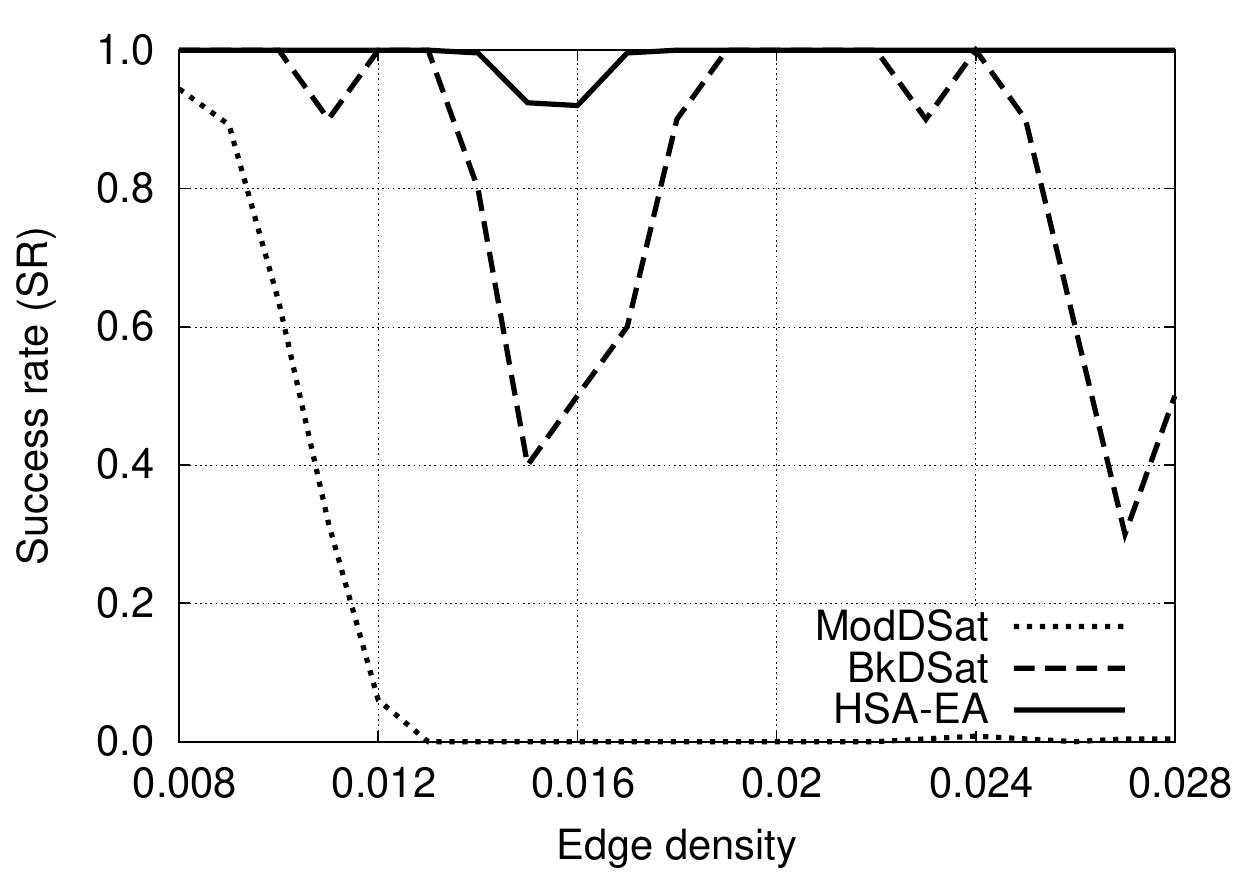}
\vspace{-5mm}
\end{SCfigure}

\begin{SCfigure}
\centering
\label{fig:Fig_MED3}
\caption{SR on flat medium-scale graphs}
\includegraphics[width=0.7\textwidth]{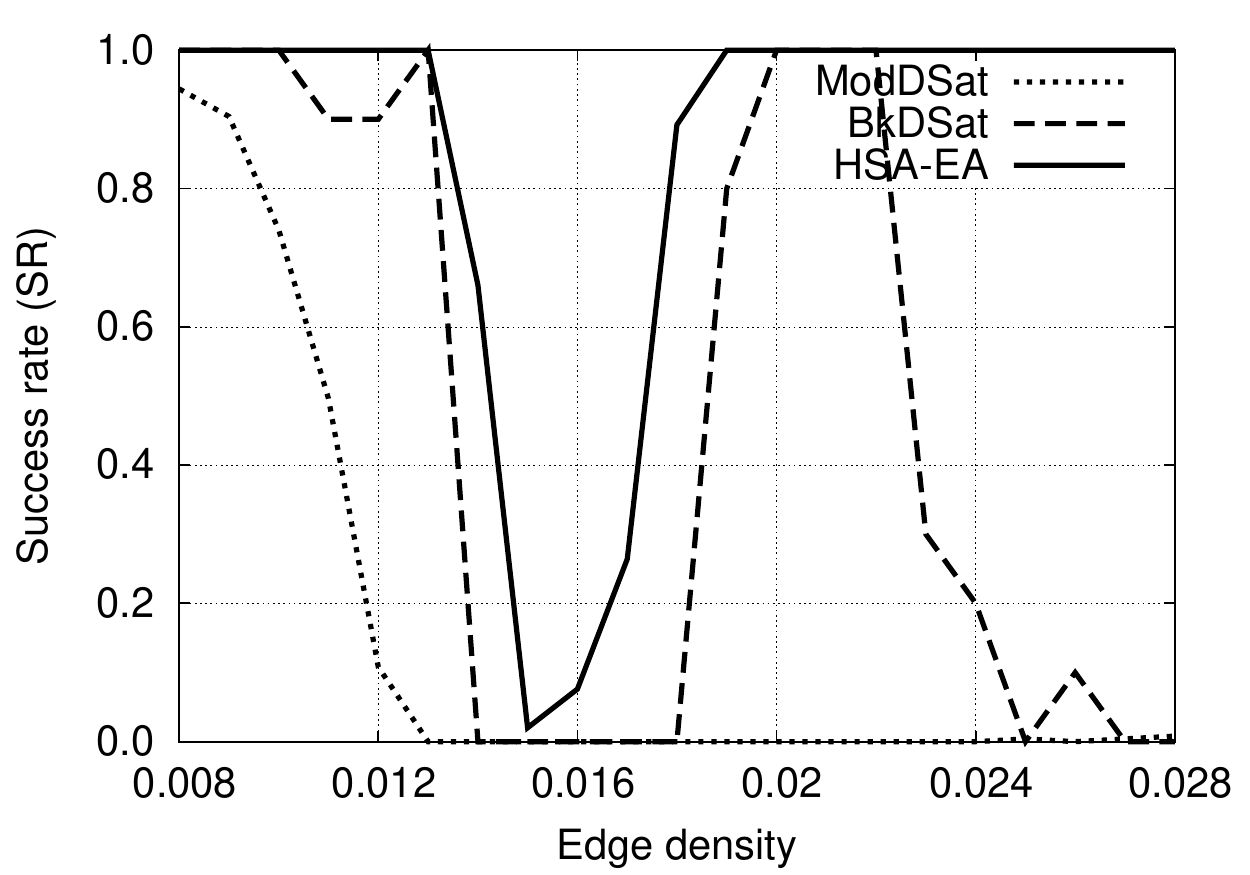}
\vspace{-5mm}
\end{SCfigure}

The medium-scale graphs were generated as described in Section 4.2.1. The ModDSat algorithm was executed 500 times by varying the first
vertex from 1 to 500. The BkDSat algorithm terminated when the solution was found or the number of backtracking steps to the first vertex that could be colored with another color reached 300,000. The same number of objective function evaluations were also considered as the termination condition for HSA-EA. All three algorithms were run on graph instances created with random seeds from 1 to 10. Each algorithm was executed 25 times on each graph instance. The results are summarized in Figs.~\ref{fig:Fig_MED1}-\ref{fig:Fig_MED3}. Note that the results are compared with regard to the SR measure only.

The behavior of both DSatur variants is similar on the graphs of all types. While BkDSat obtained $\textnormal{SR}>0$ on uniform
(Fig.~\ref{fig:Fig_MED1}) and equi-partite (Fig.~\ref{fig:Fig_MED2}) graphs, and $\textnormal{SR}=0$ on flat graphs (Fig.~\ref{fig:Fig_MED3}) in the phase transition ($p\in [0.014,0.016]$), this was not the case with ModDSat that could only successfully solve the instances with $p<0.012$.

\begin{SCfigure}
\centering
\label{fig:Fig_LAR1}
\caption{SR on uniform large-scale graphs}
\includegraphics[width=0.7\textwidth]{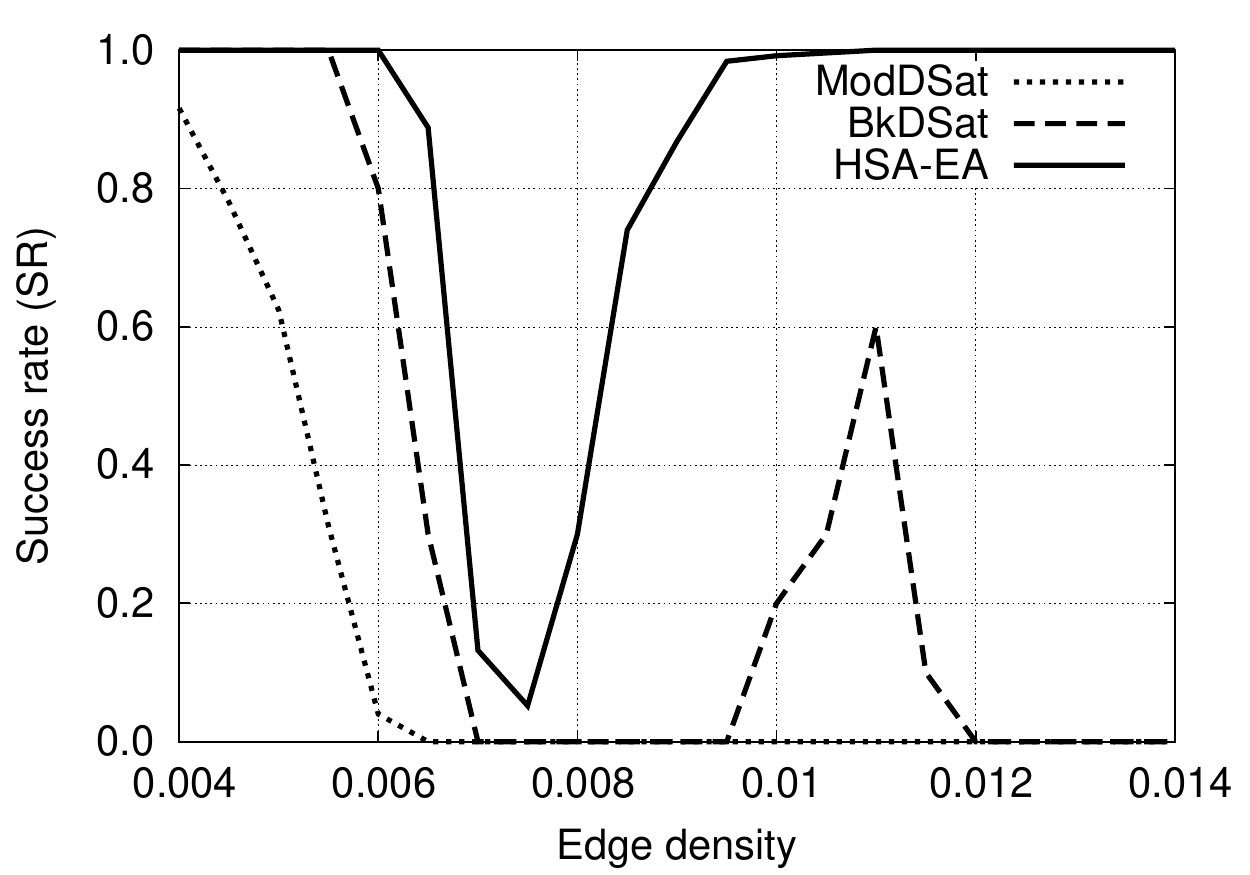}
\vspace{-5mm}
\end{SCfigure}

\begin{SCfigure}
\centering
\label{fig:Fig_LAR2}
\caption{SR on equi-partite large-scale graphs}
\includegraphics[width=0.7\textwidth]{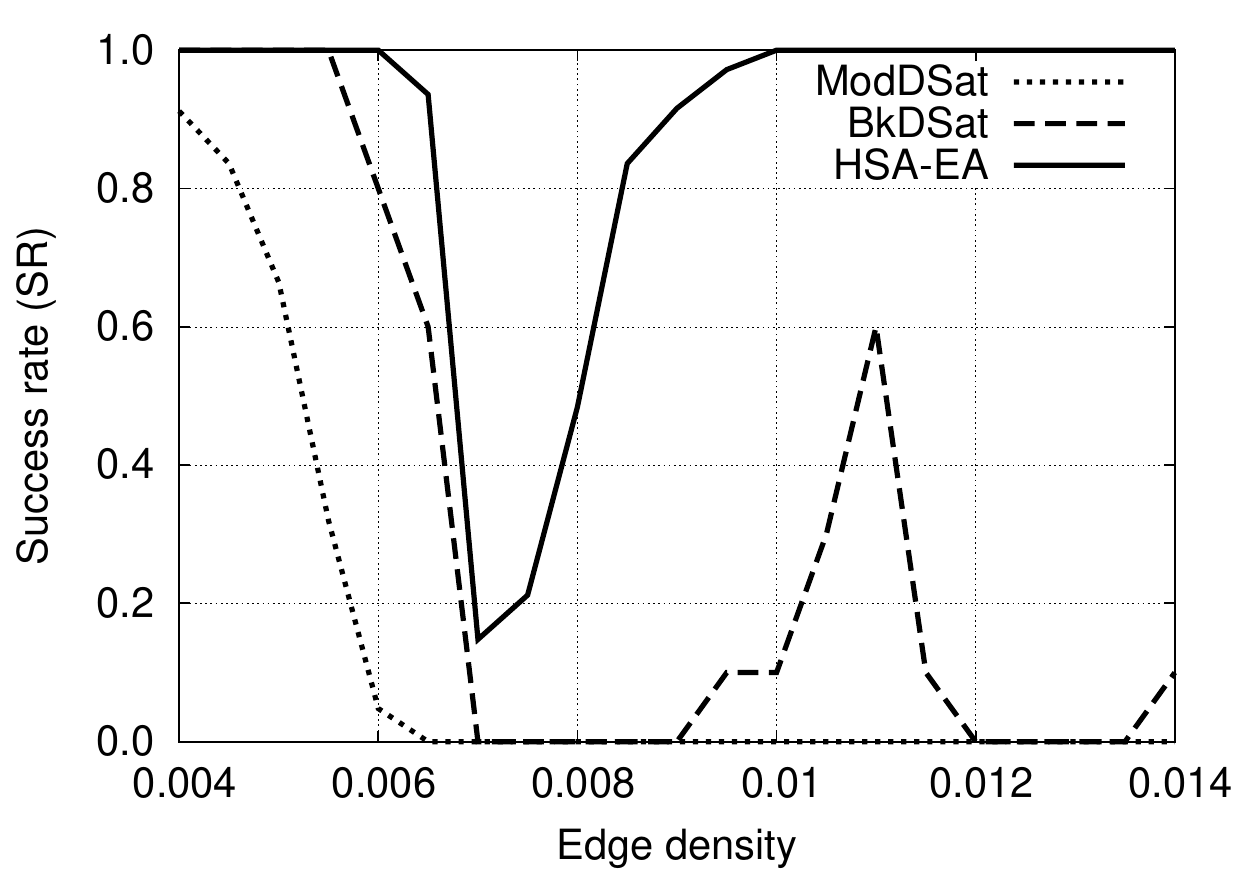}
\vspace{-5mm}
\end{SCfigure}

\begin{SCfigure}
\centering
\label{fig:Fig_LAR3}
\caption{SR on flat large-scale graphs}
\includegraphics[width=0.7\textwidth]{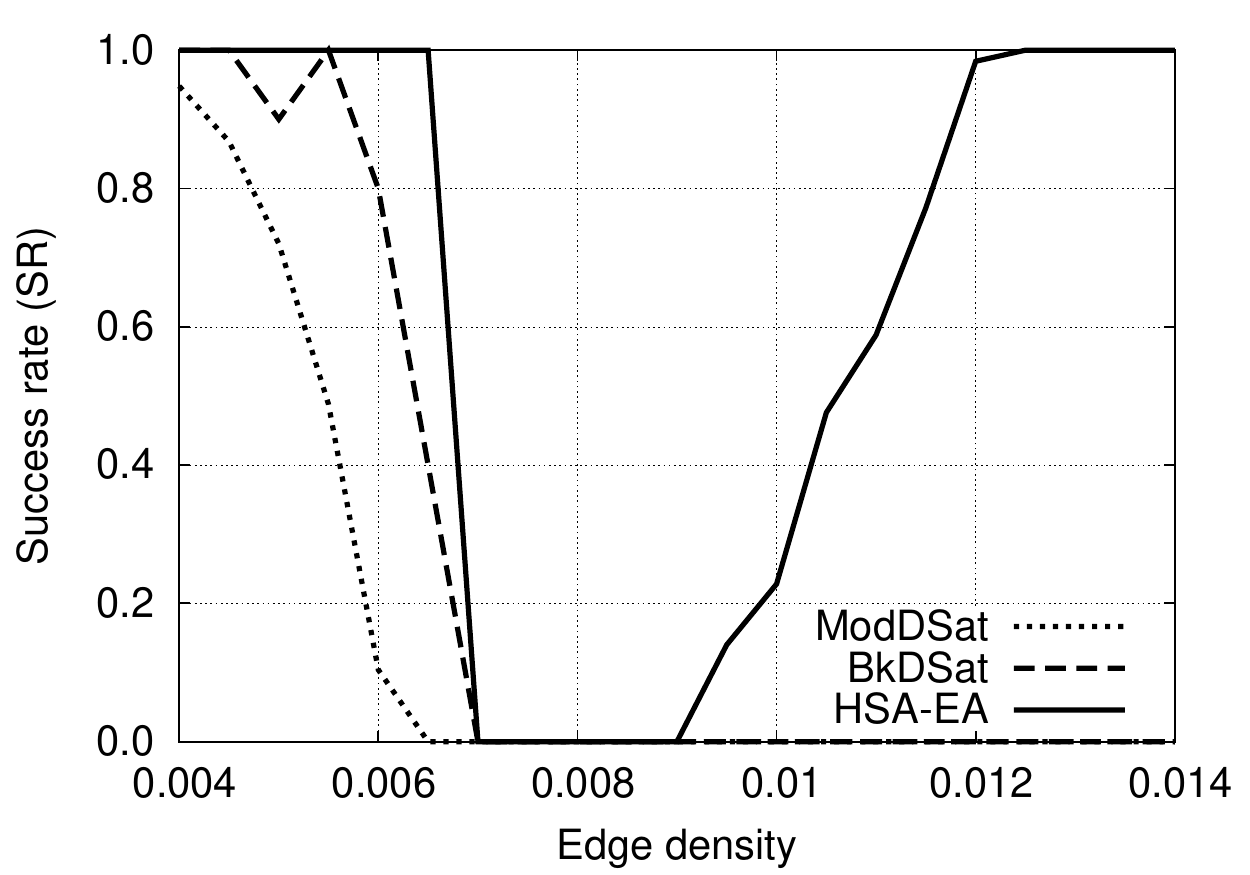}
\vspace{-5mm}
\end{SCfigure}

\subsubsection{Large-scale graphs}

Large-scale graphs were generated according to the scenario described in Section 4.2.2. However, ModDSat was executed 1,000
times. BkDSat terminated when the number of backtracking steps to the first vertex that could be colored with another color reached 300,000, while for HSA-EA when the number of objective function evaluations reached the same value. All three algorithms run graph 3-coloring created with random seeds from 1 to 10 and repeated the graph 3-coloring 25 times using different seeds. The results were accumulated for each graph instance. They are presented in Figs.~\ref{fig:Fig_LAR1}-\ref{fig:Fig_LAR3}.

The results on large-scale graphs were similar to the results on medium-scale graphs. That is, ModDSat and BkDSat successfully
3-colored the graphs below the phase transition ($p<0.007$). BkDSat 3-colored some instances of the uniform (Fig.~\ref{fig:Fig_LAR1}) and equi-partite graphs (Fig.~\ref{fig:Fig_LAR2}) in the interval $p\in[0.0095,0.012]$, while the results are poor outside this interval. On flat graphs no results were obtained for $p>0.012$. The behavior of ModDSat was even worse because it only found solutions for the graph instances with $p<0.0065$.

\subsection{Analysis of the graph structural features}

The graph size has the major impact on the performance of the graph-coloring algorithms. This is expressed as the number of vertices.
Obviously, the more vertices in the graph, the harder the graph to color. However, given a fixed graph size, additional variables
determine the hardness of the graph to color, as follows:
\begin{itemize}
  \item the graph type,
  \item the edge density and
  \item the variability in sizes of the color classes.
\end{itemize}
\noindent These variables determine the structural features of graphs and are also referred to as stratification variables.

An additional quality measure, \textit{Error rate} (ER), was defined to facilitate the analysis of the structural features. ER reflects the average number of unsuccessful runs. Obviously, the best ER is zero. This measure is derived from the success rate as follows:
\begin{equation}
\label{eq:Err}
 \textnormal{ER}=1-\overline{\textnormal{SR}},
\end{equation}
\noindent where $\overline{\textnormal{SR}}$ determines the average success rate according to the observed stratification variable.
While SR is defined as the average number of successful runs in coloring one graph instance, $\overline{\textnormal{SR}}$
also considers the average success rate over a number of instances. The next subsections present the influence of stratification
variables on the behavior of the graph-coloring algorithms.

\subsubsection{Influence of the graph size}

The impact of graph size on the performance of graph-coloring algorithms was investigated for two graph sizes, i.e., $n=500$
(medium-scale graphs) and $n=1,000$ (large-scale graphs). Here, $\overline{\textnormal{SR}}$ from Eq.~\ref{eq:Err} is defined as the
average SR for various types of medium-scale graphs varying $p \in [0.008,0.028]$ with a step of 0.001, and large-scale graphs varying
$p \in [0.004,0.014]$ with a step of 0.0005. These intervals were selected to cover the respective phase transition regions. The SRs were then aggregated for each tested algorithm.

\begin{figure}[hbt]	%
\centering
\subfigure[Graph size] {\includegraphics[width=5.8cm]{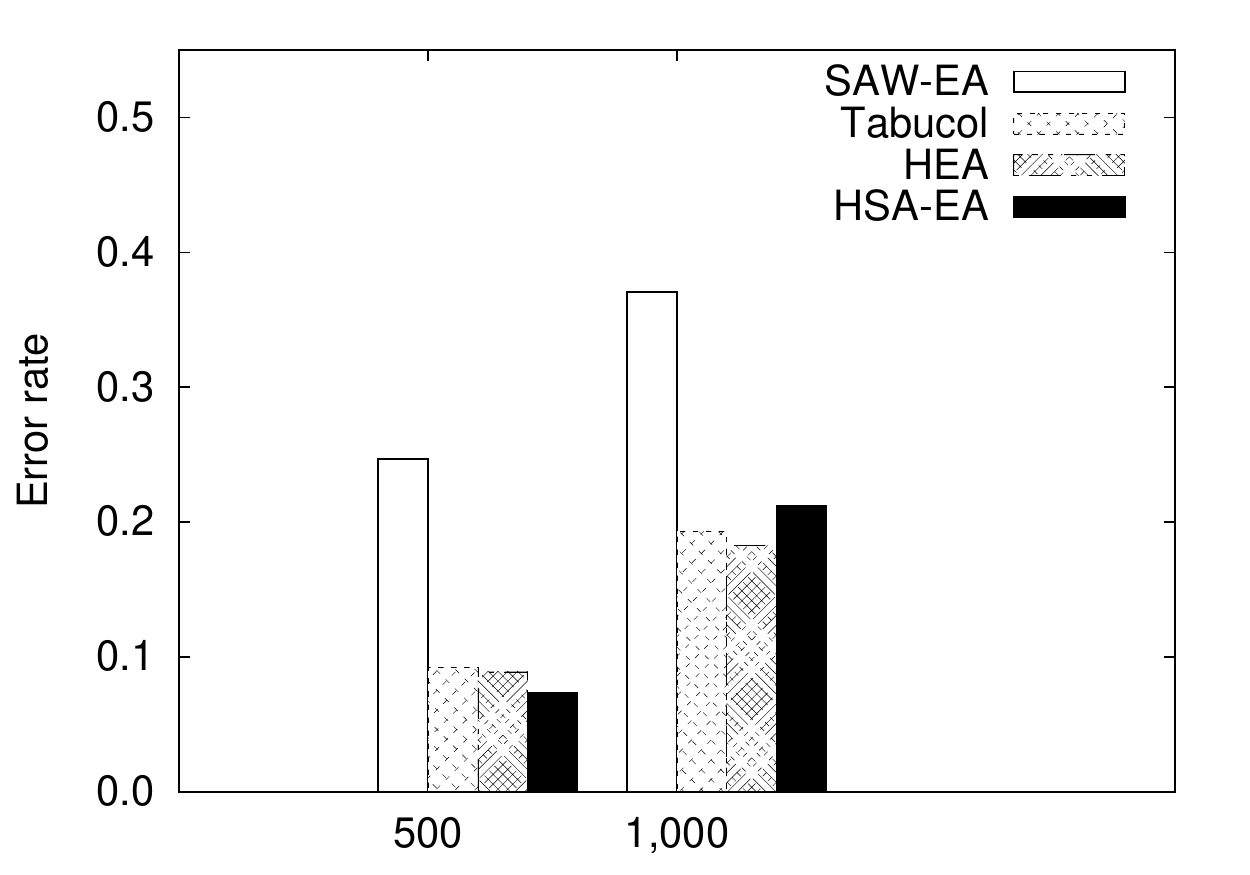}}
\subfigure[Graph type] {\includegraphics[width=5.8cm]{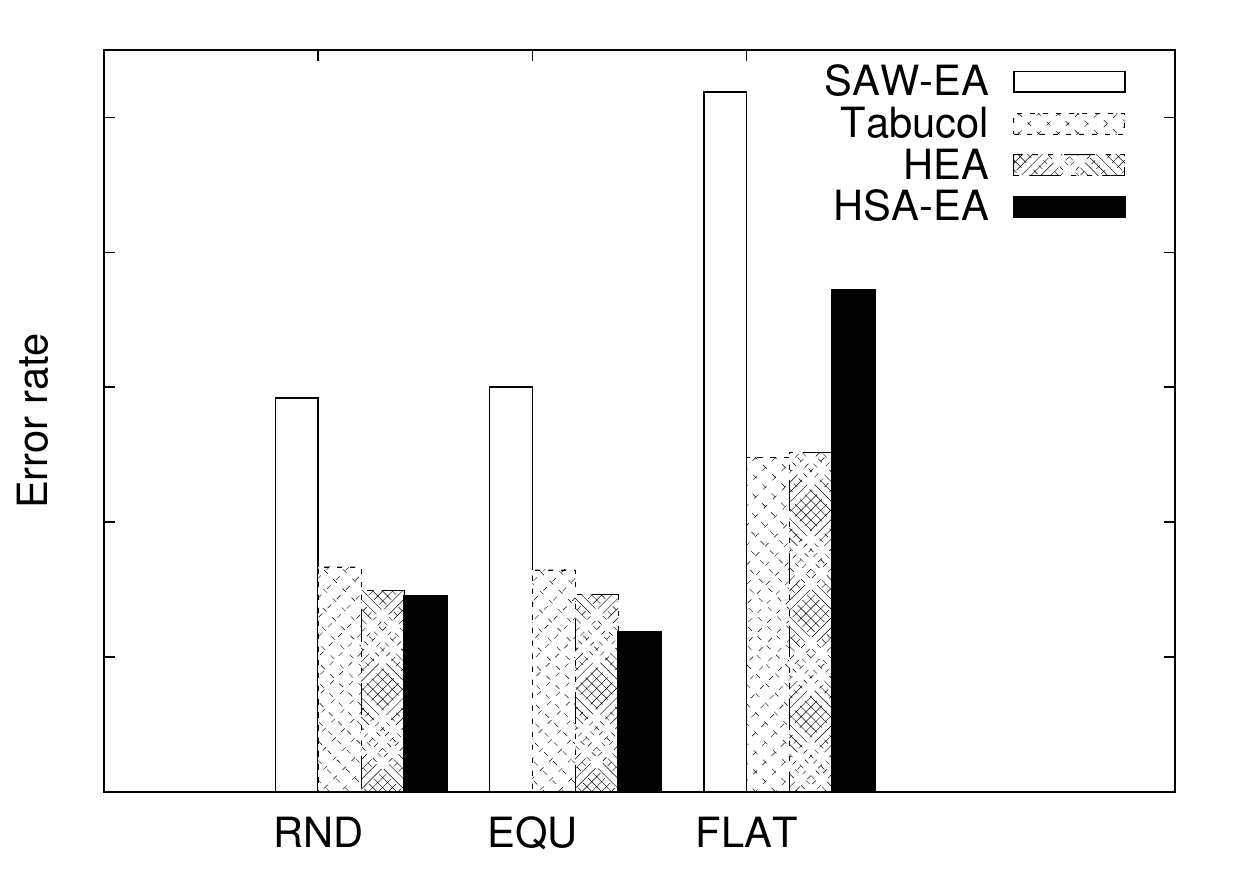}}
\caption{Influence of graph size and type on the performance of graph-coloring algorithms}
\label{fig:Sub_2}
\end{figure}

As seen from Fig.~\ref{fig:Sub_2}.a, the graph size has a big influence on the algorithm performance. The ERs for all tested algorithms increased (and SR declined) with increasing graph size. The best results on medium-scale graphs were gained by HSA-EA ($\textnormal{ER}=0.07$), while large-scale graphs were best solved by HEA ($\textnormal{ER}=0.18$). In both cases, SAW-EA achieved the worst results. Specifically, the improvement of the results as obtained by HEA is more gradual than by HSA-EA. That is, HEA is less sensitive to increasing graph size than HSA-EA.

\subsubsection{Influence of the graph type}

The performance of the graph-coloring algorithms was analyzed with regard to the graph type. The following types of graphs were taken
into consideration: uniform, equi-partite, and flat. Note that here, only the large-scale graphs were considered. In this experiment, the $\overline{\textnormal{SR}}$ from~(\ref{eq:Err}) was defined as the average SR achieved when coloring various types of large-scale graphs varying $p \in [0.004,0.014]$ with a step of 0.0005.

As seen from Fig.~\ref{fig:Sub_2}.b, the best result ($\textnormal{ER}=0.12$) was achieved by HSA-EA on equi-partite graphs. In general, there existed only a small difference between coloring of uniform and equi-partite graphs when comparing particular algorithms, while the flat graphs were the hardest to all tested algorithms. When comparing the performance of individual algorithms, it could be seen that the best results were obtained by HSA-EA, which outperformed the HEA on the uniform and equi-partite graphs. Tabucol was slightly worse, but SAW-EA did not reach the performance of the other algorithms. The flat graphs were best colored by Tabucol and HEA, slightly worse by HSA-EA, and worst by SAW-EA.

\begin{figure}[hbt]	
\centering
\subfigure[Edge density] {\includegraphics[width=5.8cm]{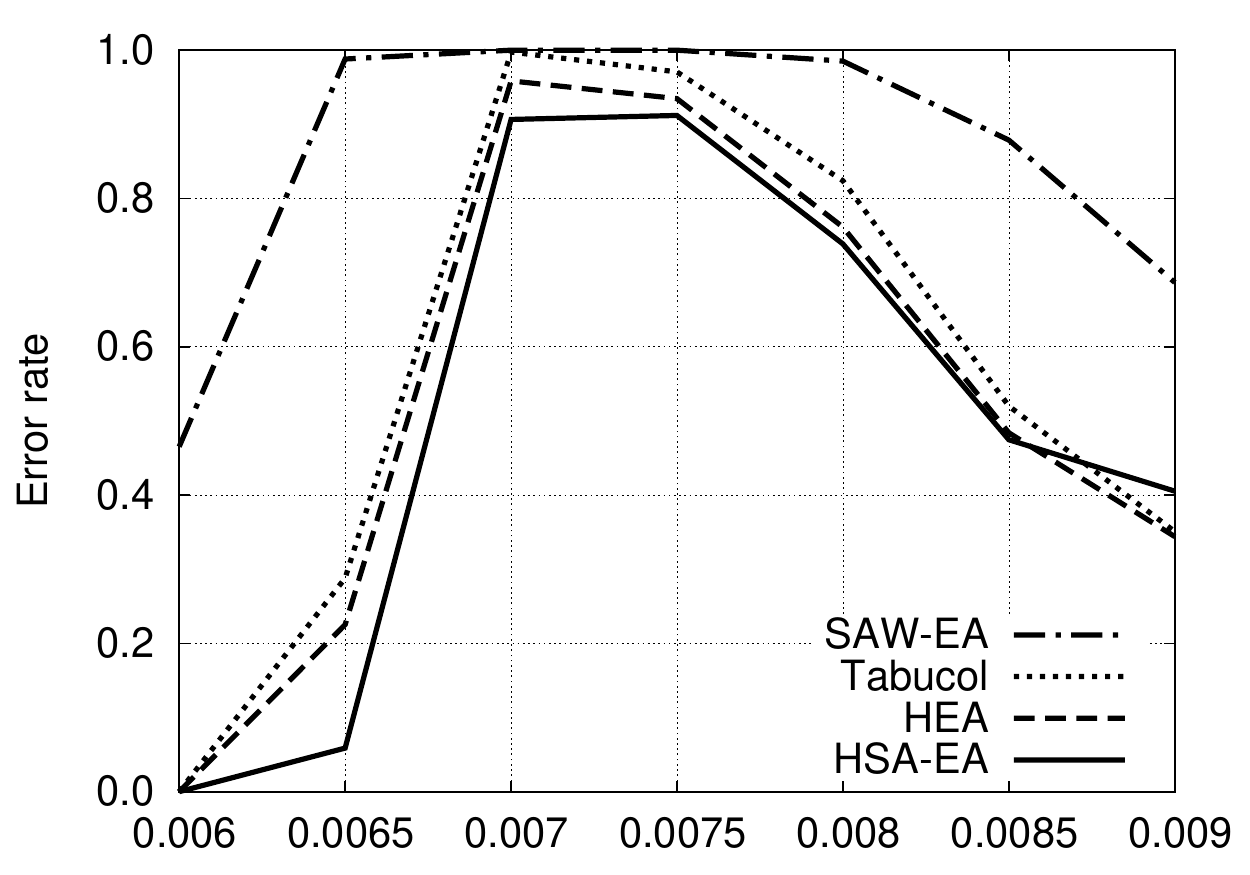}}
\subfigure[Variability in sizes of the color classes] {\includegraphics[width=5.8cm]{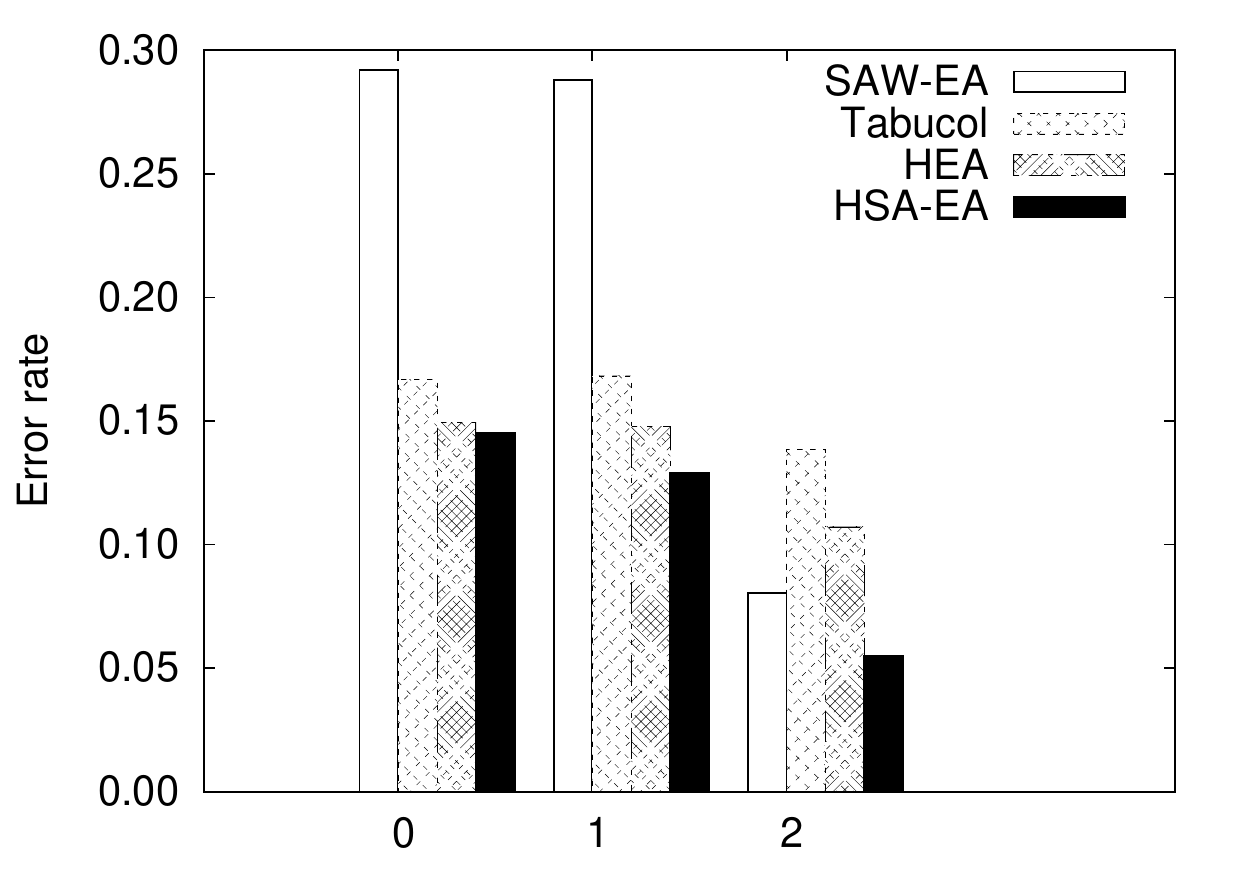}}
\caption{Influence of the edge density and the variability in sizes of the color classes on the performance of graph-coloring
algorithms}
\label{fig:Sub_3}
\end{figure}

\subsubsection{Influence of the edge density}

The influence of the edge density when coloring the large-scale graphs has already been discussed in Section~\ref{sub_sec:large}, where the results were evaluated according to SR. Here, the interest is to investigate the influence of edge density on the performance of the graph-coloring algorithms. Therefore, the ER results from coloring the three types of large-scale graphs varying $p\in[0.006,0.009]$
with a step of 0.0005 were aggregated for each coloring algorithm. Note that the observed interval of the edge density precisely
corresponded to the phase transition region.

As illustrated in~Fig.~\ref{fig:Sub_3}.a, the best results were achieved by HSA-EA that colored all instances of the observed graphs. HEA and Tabucol were also close to this result. Actually, graph 3-coloring in the hardest instance with $p=0.007$ was best performed by HSA-EA, followed by HEA and Tabucol, while graph instances in the phase transition, i.e., at $p=0.007$ and at $p=0.0075$, remained a challenge for SAW-EA.

\subsubsection{Influence of the variability in sizes of the color classes}

This influence was studied to confirm the assertion of Turner~\cite{Turner:1988} that the variable 3-colorable graphs are easier to
color. The results of the tested graph-coloring algorithms were compared in 3-coloring the large-scale graphs varying $p
\in [0.004,0.014]$ with a step of 0.0005.

As shown in Fig.~\ref{fig:Sub_3}.b, the results confirmed that the assertion by Turner holds for all graph-coloring algorithms. The variability in sizes of the color classes had the highest impact on the performance of SAW-EA that 3-colored the graphs with $\Delta = 2$ even better than HEA and Tabucol. Unfortunately, the results of this algorithm were much worse when compared with the results of the other algorithms on graphs with variabilities $\Delta = 0$ and $\Delta = 1$. According to this parameter, HSA-EA generally outperformed all other algorithms.

\subsection{Statistical analysis of results}

To evaluate the quality of the graph-coloring algorithms, a non-parametric analysis of their results was carried out, as suggested by
Dem\v{s}ar~\cite{Demsar:2006}. As a basis, the Friedman non-parametric test was considered~\cite{Friedman:1937,Friedman:1940}. This test compares the average ranks of algorithms. A null-hypothesis states that two algorithms are equivalent and, therefore, their ranks should be equal. If the null-hypothesis is rejected, i.e., the performance of the algorithms is statistically different, the Bonferroni-Dunn test~\cite{Demsar:2006} is performed that calculates the critical difference between the average ranks of those two algorithms. When the statistical difference is higher than the critical difference, the algorithms are significantly different. The equation for the calculation of critical difference can be found in~\cite{Demsar:2006}.

The Friedman non-parametric tests refered to the large-scale graphs only. Two Friedman non-parametric tests were performed. In the first test, the behavior of algorithms was observed when coloring specific instances of different graph types in the phase transition. For this purpose, 30 uniform graphs, 10 equi-partite, and 10 flat graphs in the phase transition were taken into account. In the second test, the average results of the graph 3-coloring algorithms were compared in the phase transition. In this case, the edge densities $p \in [ 0.006, 0.009 ]$ with a step of 0.0005 were treated, obtaining 7 graph instances. Moreover, the average results on all uniform graphs with different variabilities in sizes of the color classes $\Delta \in \{0,1,2\}$ were taken during this analysis and, in line with this, the number of graph instances increased to 350, i.e., $5 \times 10 \times 7$.

\begin{figure}[hbt]	
\subfigure {\includegraphics[width=3.8cm]{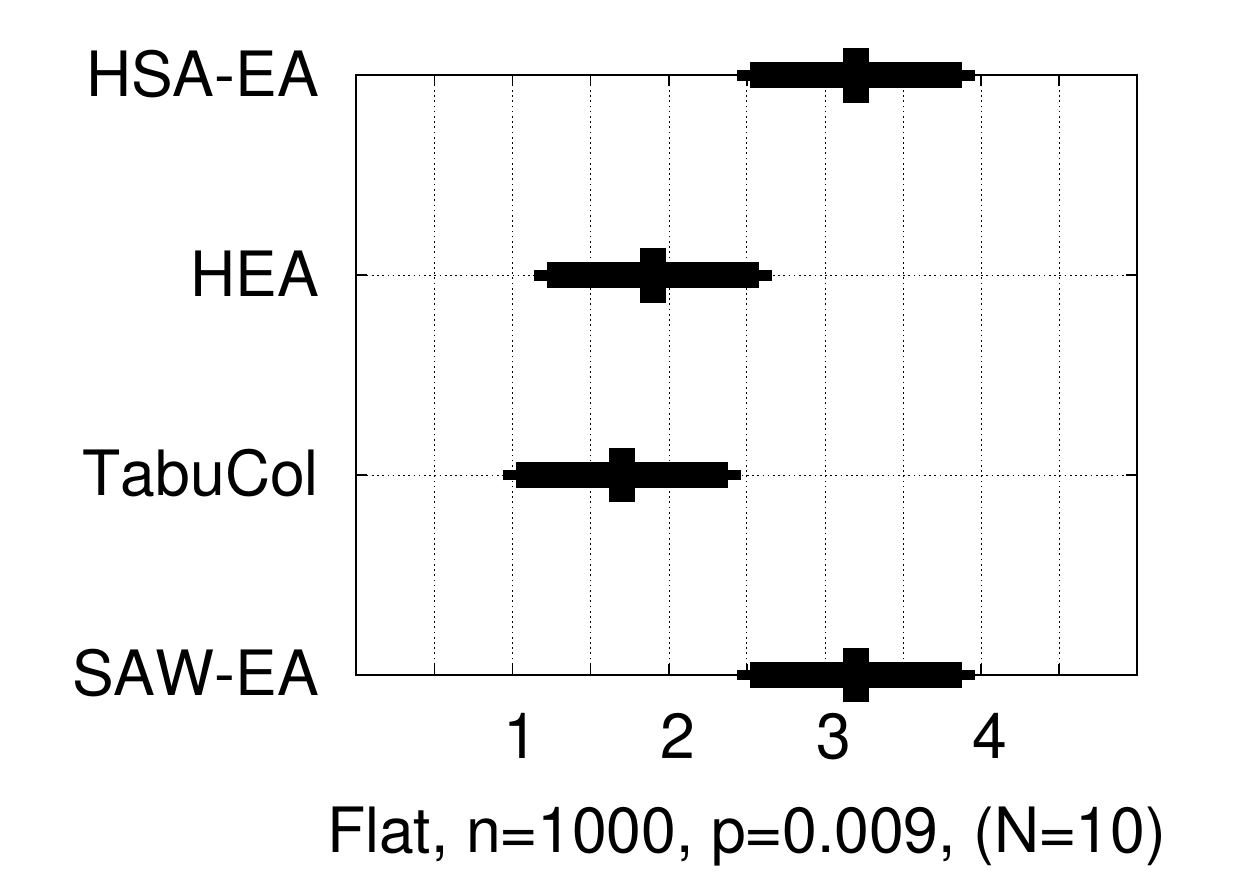}}
\subfigure {\includegraphics[width=3.8cm]{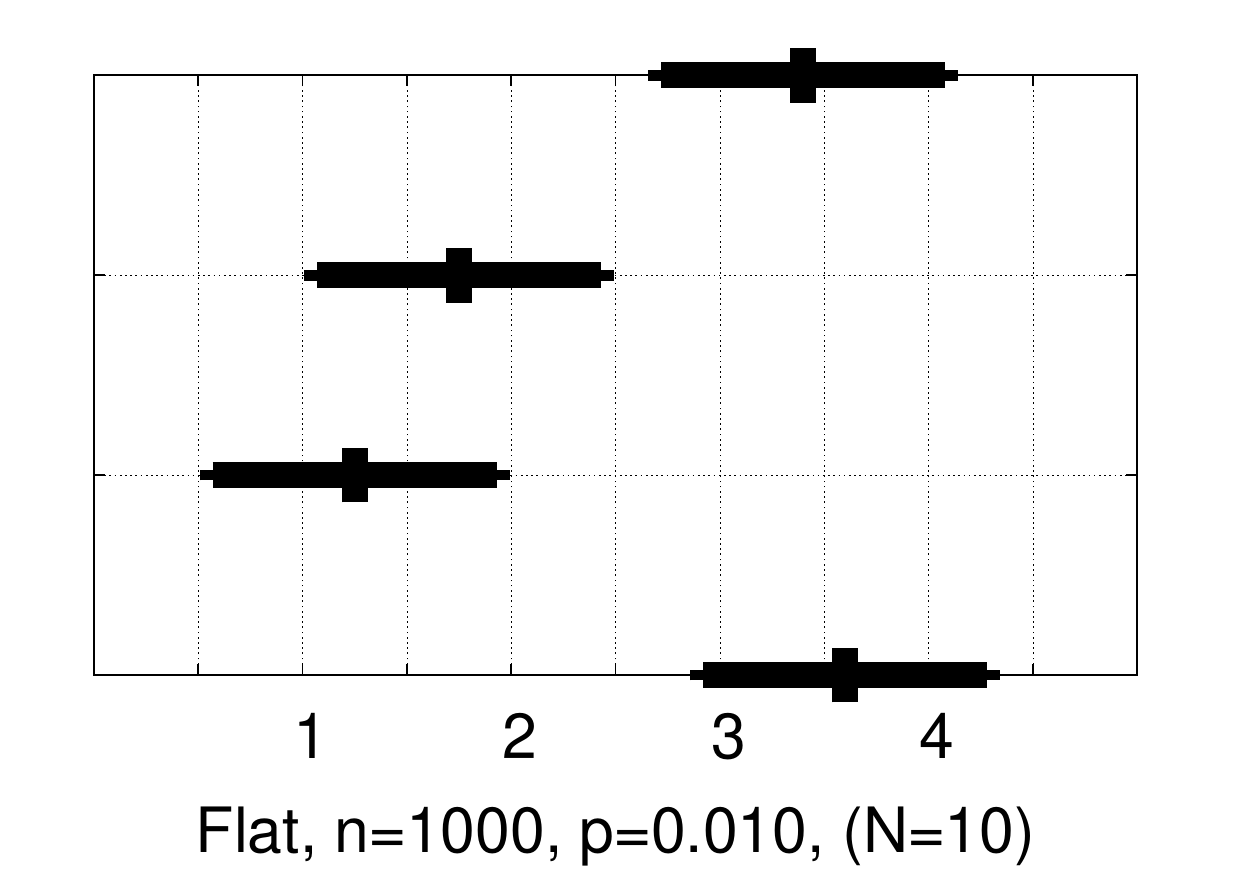}}
\subfigure {\includegraphics[width=3.8cm]{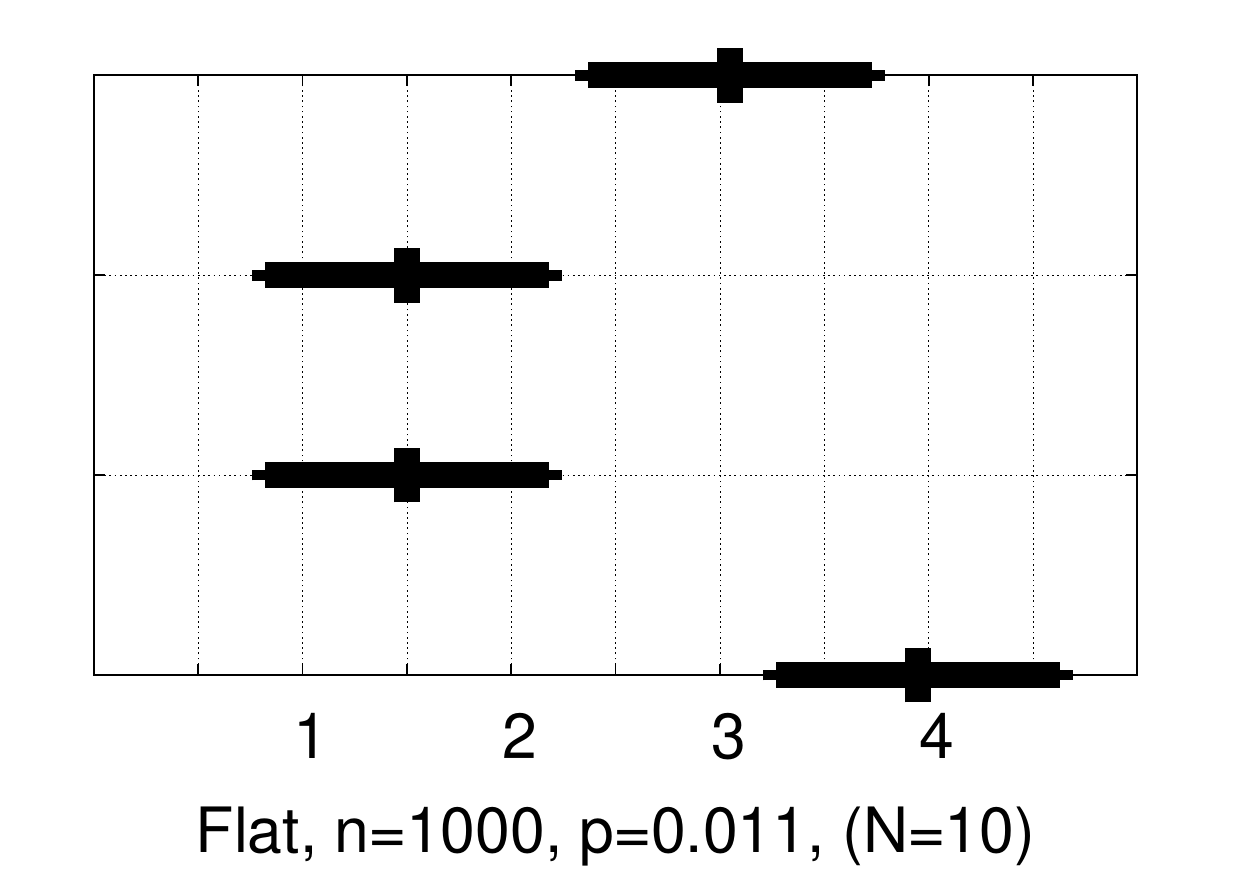}}
\caption{Results of the Friedman non-parametric test on the specific large-scale graph instances}
\label{fig:Sub_5}
\end{figure}

The results of the first Friedman non-parametric test are presented in Fig.~\ref{fig:Sub_5} being divided into nine diagrams that show the ranks and confidence intervals (critical differences) for the algorithms under consideration. The diagrams are organized according to the graph types and the edge densities. Two algorithms are significantly different if their intervals in~Fig.~\ref{fig:Sub_5} do not overlap. Note that the uniform and equi-partite graphs with edge densities $p=\{0.007,0.008,0.009\}$ are considered, while the flat graphs are taken with edge densities $p=\{0.009,0.010,0.011\}$ because the tested algorithms did not find any solution coloring the graphs of this type with edge densities $p=\{0.007,0.008\}$.

The following conclusions can be derived from Fig.~\ref{fig:Sub_5}:
\begin{itemize}
  \item On uniform graphs, HSA-EA significantly improves the results of Tabucol and SAW-EA on instances with $p=0.007$, while HSA-EA and HEA are significantly better than Tabucol on instances with $p=0.008$. In the case of instances with $p=0.009$, all other algorithms are significantly better than SAW-EA.
  \item On equi-partite graphs, no significant difference can be detected on graphs with $p=0.007$, while all other algorithms color graphs with $p=0.008$ and $p=0.009$ significantly better than SAW-EA.
  \item On flat graphs, no significant difference can be detected on graphs with $p=0.007$, while coloring of other graph instances is performed significantly better by HEA and Tabucol than HSA-EA and SAW-EA.
\end{itemize}

\begin{figure}[hbt]	
\centering
\subfigure {\includegraphics[width=3.8cm]{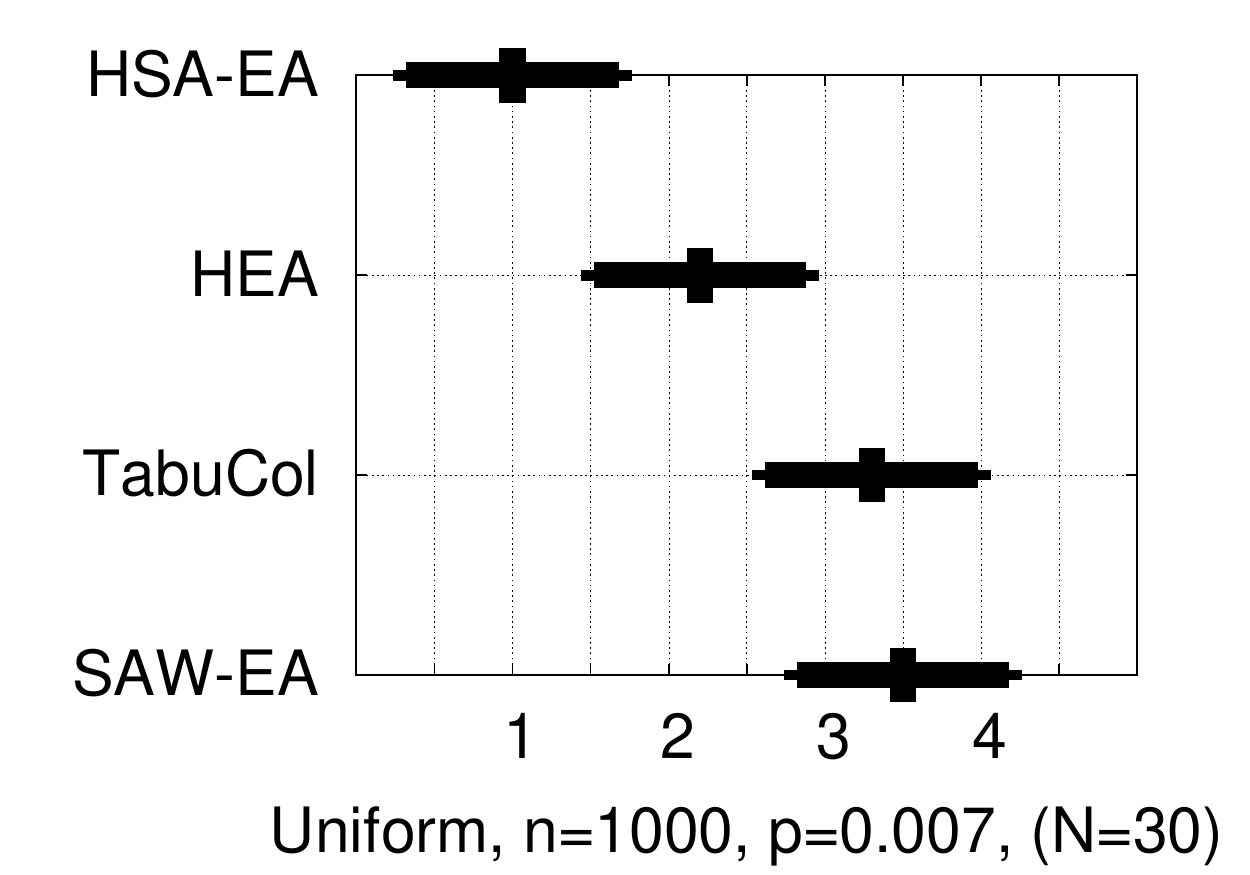}}
\subfigure {\includegraphics[width=3.8cm]{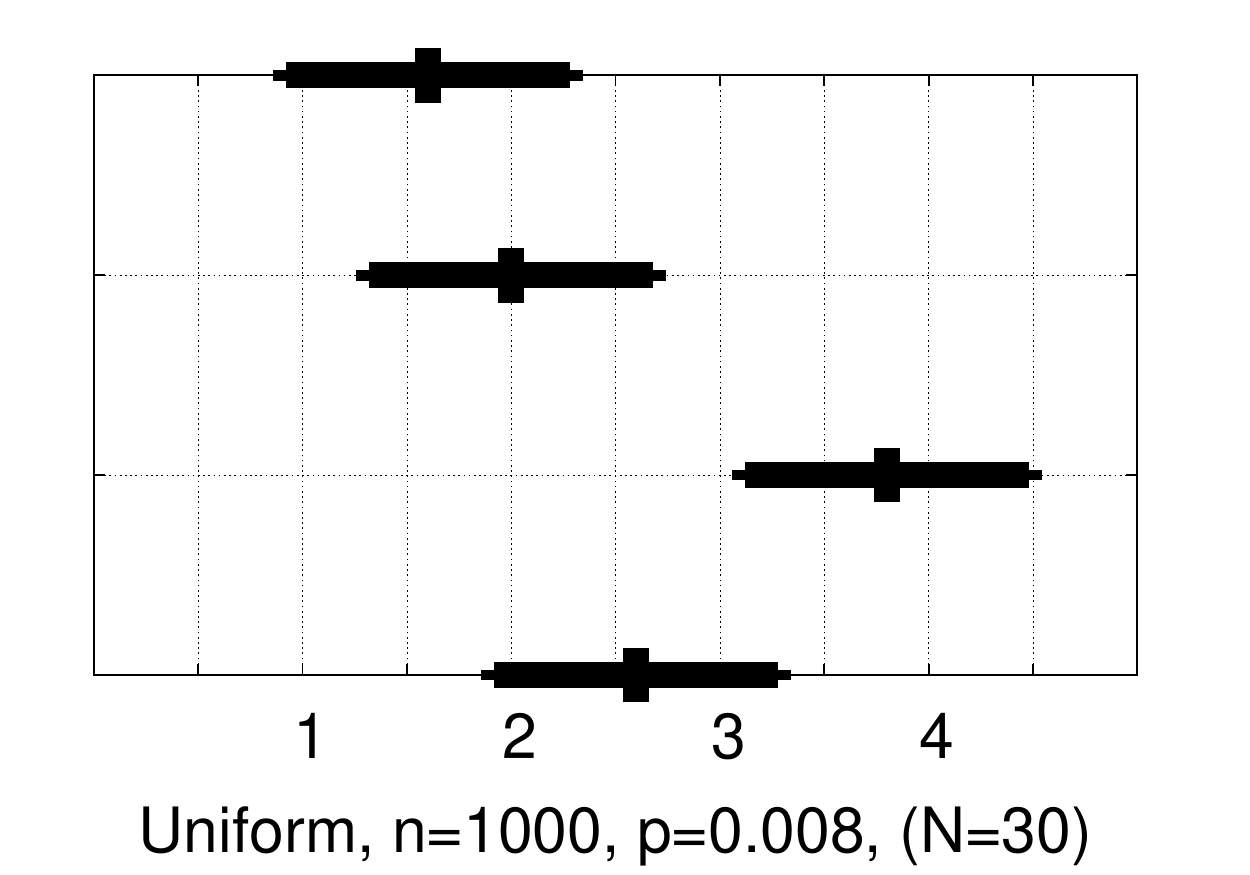}}
\subfigure {\includegraphics[width=3.8cm]{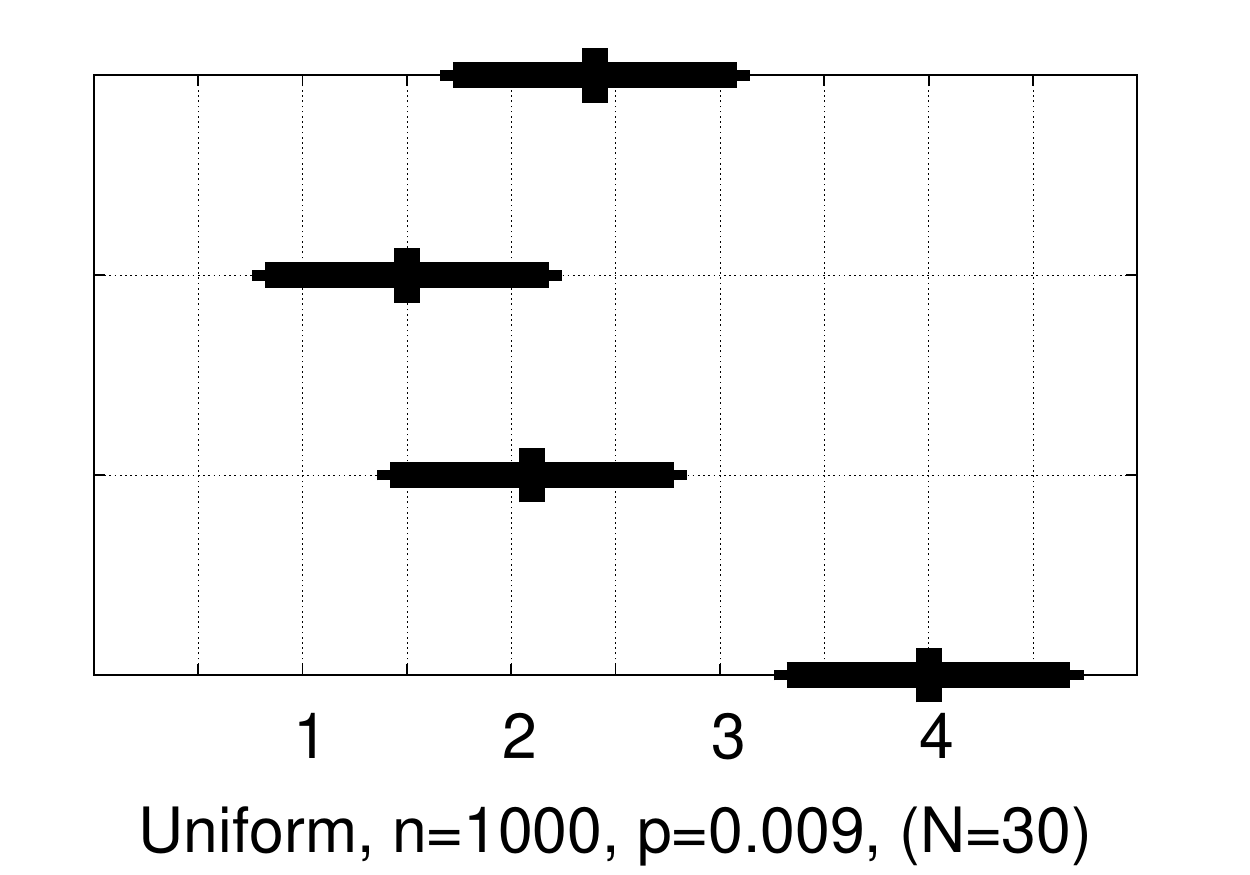}}
\end{figure}
\begin{figure}[hbt]	

\subfigure {\includegraphics[width=3.8cm]{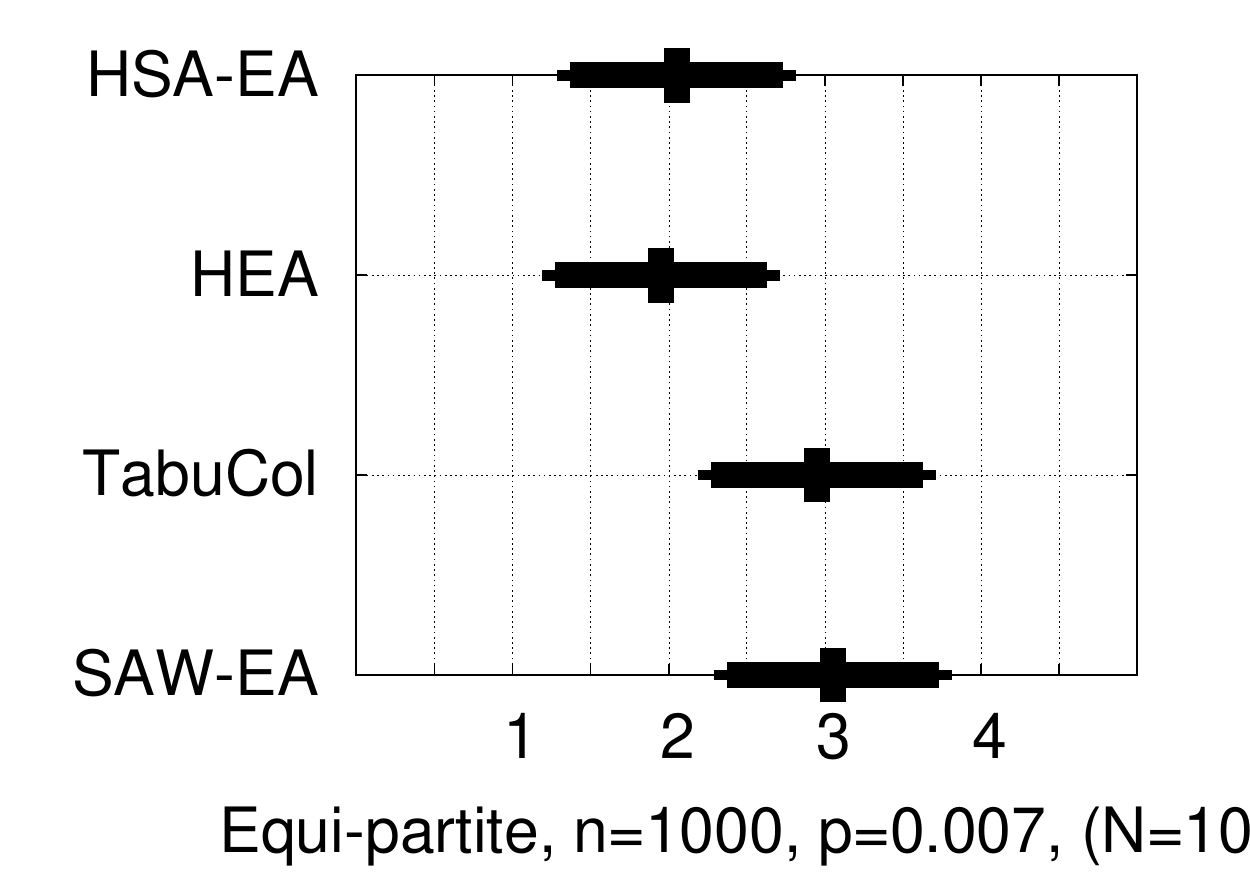}}
\subfigure {\includegraphics[width=3.8cm]{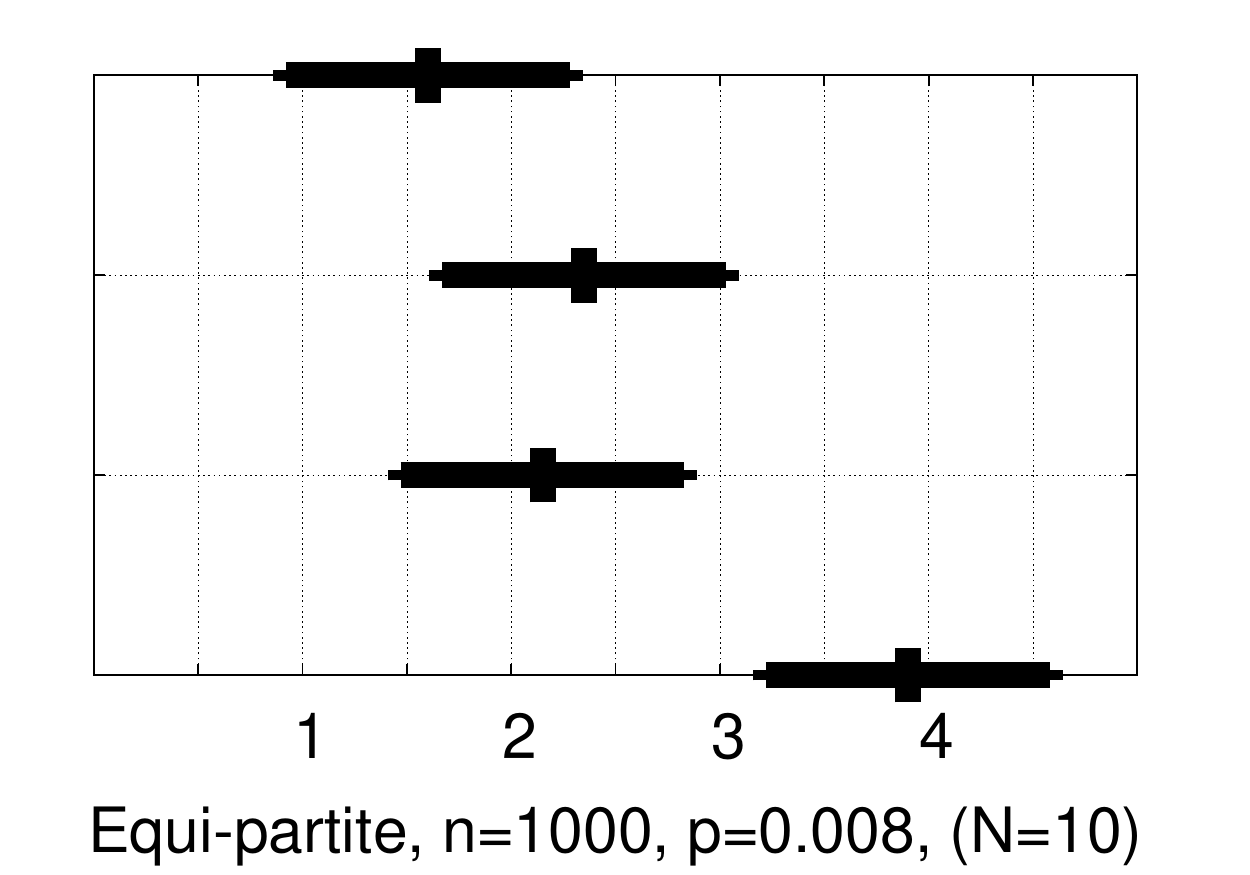}}
\subfigure {\includegraphics[width=3.8cm]{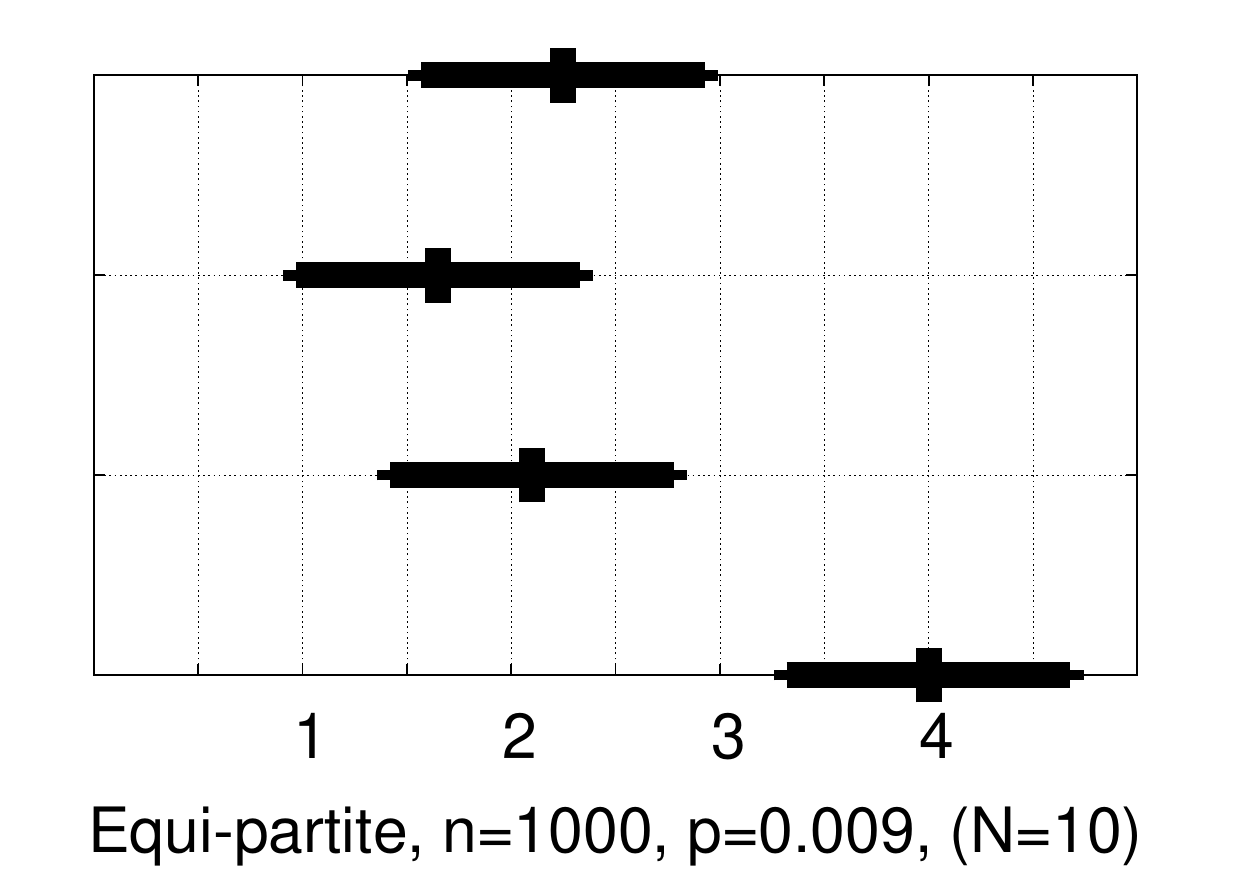}}
\end{figure}

\begin{figure}[hbt]	
\centering
\subfigure[Uniform] {\includegraphics[width=3.8cm]{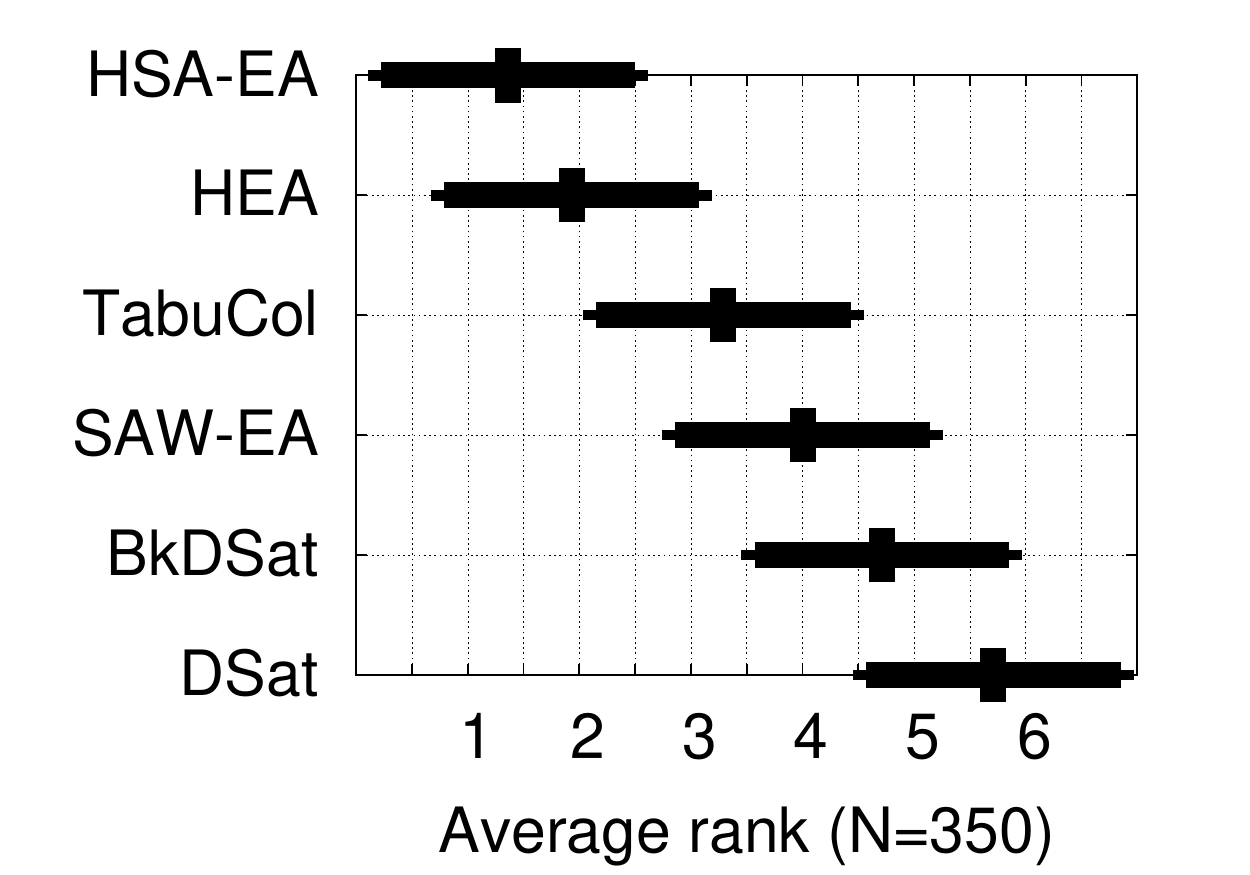}}
\subfigure[Equi-partite] {\includegraphics[width=3.8cm]{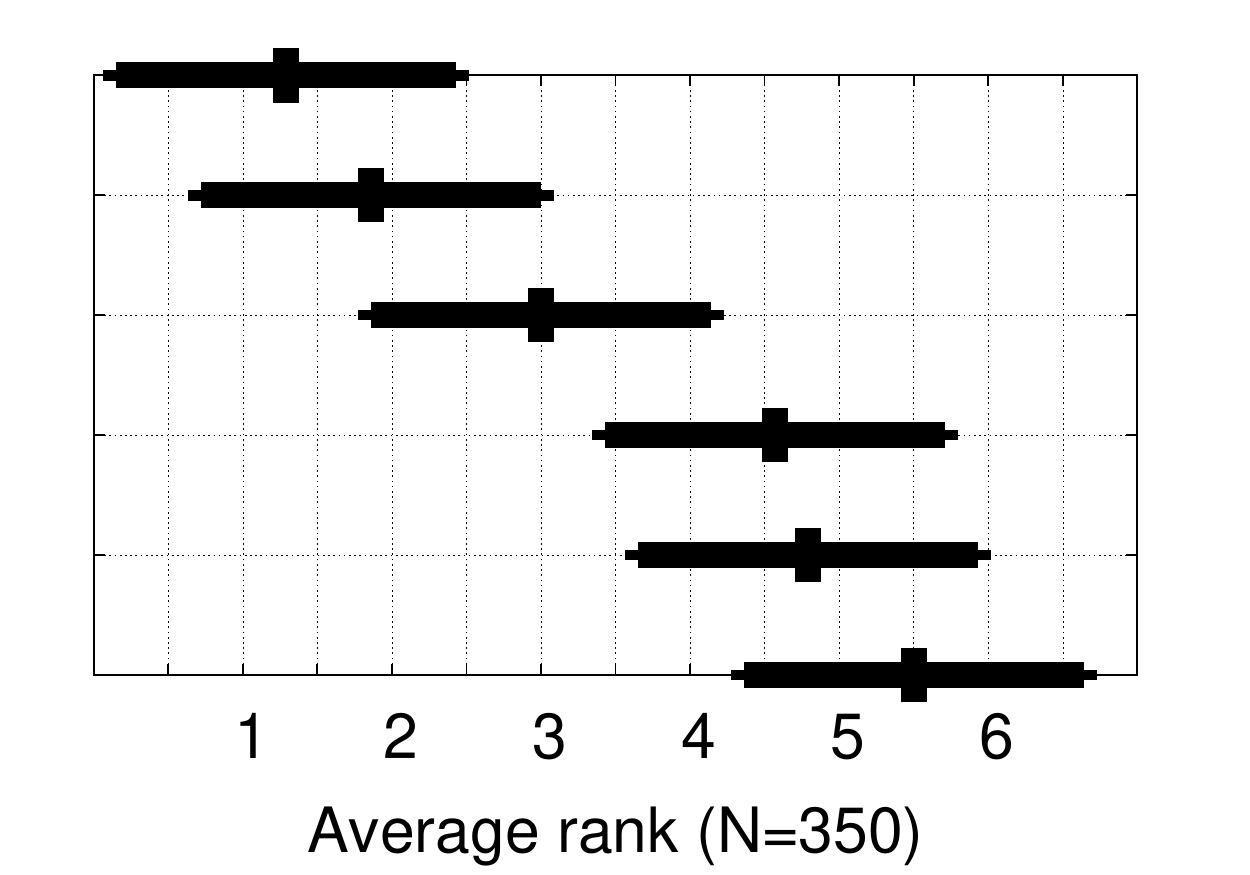}}
\subfigure[Flat] {\includegraphics[width=3.8cm]{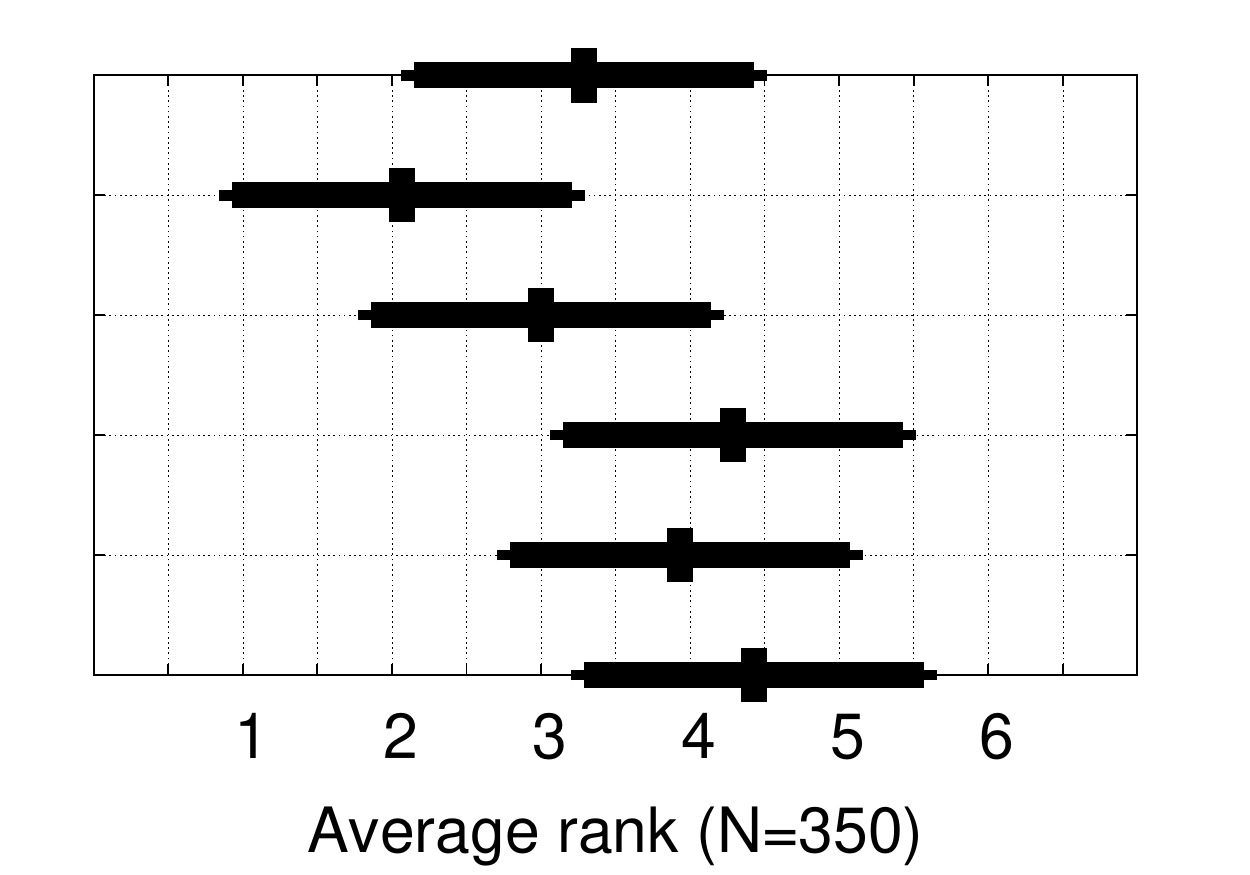}}
\caption{Averaged results of the Friedman non-parametric test on large-scale graphs}
\label{fig:Sub_4}
\end{figure}

The results of the second Friedman non-parametric test are presented in Fig.~\ref{fig:Sub_4} showing the average ranks and confidence intervals for the tested algorithms. Here, besides the algorithms mentioned in the first test, both variants of the DSatur algorithm were also observed. The figure consists of three diagrams that correspond to three graph types, i.e., uniform, equi-partite, and flat. Two algorithms are significantly different if their intervals in~Fig.~\ref{fig:Sub_4} do not overlap.

The following conclusions can be inferred from the results of the Friedman tests:
\begin{itemize}
  \item On uniform graphs, HSA-EA and HEA significantly improve the results of BkDSat and ModDSat. The results of
      HSA-EA are slightly better than the results of HEA.
  \item On equi-partite graphs, the results of HSA-EA and HEA significantly improve the results of SAW-EA, BkDSat, and ModDSat. Essentially, HSA-EA improves the results of HEA, but the difference is not significant.
  \item On flat graphs, the results of the tested algorithms are comparable.
\end{itemize}

It should to be noted that all the compared algorithms are tailored to the problem in a specific manner. For example, the DSatur algorithm is a traditional heuristic for graph coloring that sequentially colors the graph vertices according to the saturation degrees. This algorithm can be improved with backtracking that exploits the tie breaking. Moreover, HSA-EA represents a hybridization of the evolutionary algorithm with the original DSatur algorithm. Moreover, Tabucol is a well known traditional heuristic for graph coloring. As matter of fact, the application of this heuristic to the evolutionary algorithm has led to HEA that improved the results of the original Tabucol. Finally, the SAW mechanism tries to improve the behavior of the
classical evolutionary algorithm in graph coloring.

By analyzing how successfully domain-specific knowledge can be incorporated into the particular evolutionary algorithm, it can be concluded that hybridization of the evolutionary algorithm with the DSatur heuristic (HSA-EA) exhibited better results on uniform and equi-partite graphs, while coloring flat graphs was better performed by the evolutionary algorithm hybridized with the Tabucol heuristic (HEA). On the other hand, applying the SAW mechanism improves the results of the original evolutionary algorithm, but this improvement does not outperform HSA-EA and HEA.

\section{Conclusions}

In this paper, we have proposed the HSA-EA for graph 3-coloring, which is hybridized with hybrid genotype-phenotype mapping, a swap local search heuristic, and the neutral survivor selection operator. In addition, a heuristic initialization procedure is applied by HSA-EA. This evolutionary algorithm was compared with SAW-EA~\cite{Eiben:1998}, Tabucol~\cite{Hertz:1987}, and HEA~\cite{Galinier:1999}.

The objective of our experiments was threefold: first, to investigate the behavior of the tested graph-coloring algorithms in
the phase transition, where the graph instances are hard to color, second, to indicate the impact of hybridizing the evolutionary algorithm with the DSatur traditional heuristic and third, to analyze how various structural features of randomly generated graphs influence the performance of the graph-coloring algorithms.

To satisfy the first objective, the graphs were generated varying the probability $p \in [0.008,0.028]$ with a step of 0.001 for medium-scale graphs ($n=500$) and $p \in [0.004,0.014]$ with a step of 0.0005 for large-scale graphs ($n=1,000$). In both cases, 21
instances of random 3-colorable graphs of three types were obtained, i.e., uniform, equi-partite and flat. However, all
the tested algorithms encountered troubles when coloring graph instances in the phase transition region.

To satisfy the second objective, two modified versions of the DSatur algorithm were taken into account. The first one (ModDSat) was inspired by the sequential coloring of the original DSatur algorithm that seems to be dependent on the selection of the first vertex. The second one (BkDSat) was the standard backtracking DSatur algorithm~\cite{website:Trick}. Both DSatur versions showed poor performance when coloring the medium- and large-scale graphs, not only in the phase transition but also outside this region, i.e., at $p > 0.028$ for medium-scale graphs and $p > 0.014$ for large-scale graphs. On the other hand, the results of ModDSat indicate that selection of the first vertex in DSatur is not so crucial.

To satisfy the third objective, the influence of the graphs structural features, i.e., the graph size, type, edge density and variability in sizes of the color classes, on the performance of the tested graph-coloring algorithms was compared. Regarding the graph size, we can conclude that increasing the graph size decreases the performance of the graph-coloring algorithms. Actually, HEA was the least sensitive to the increasing graph size. Flat graphs represented the hardest graph type to color for all considered graph-coloring algorithms. In regard to the edge density, HSA-EA was the most successful as it solved all instances of the observed graphs. The variability in sizes of the color classes most influenced SAW-EA. In general, on uniform graphs HEA significantly improved the results of SAW-EA, while on equi-partite graphs HSA-EA was better than HEA. However, both algorithms are significantly better than SAW-EA. Unfortunately, all considered algorithms performed poorly on the flat graphs.

Our future work in graph coloring using evolutionary algorithms will focus on $k$-GCP. A key to improved performance of the evolutionary algorithm might be to use direct representation of solutions in the form of vertex color vectors instead of vertex permutations used now. Furthermore, a new local search heuristic is needed that would better improve the solutions. It is our intention to further enhance the results on the hardest large-scale graph instances of all three classes in the phase transition region.

\bigskip{\small \smallskip\noindent Updated 5 January 2012.}

\begin{thebibliography}{10}

\bibitem{Aarts:1997}
E.~Aarts and J.K. Lenstra.
\newblock {\em {L}ocal Search in Combinatorial Optimization}.
\newblock Princeton University Press, Princeton and Oxford, 1997.

\bibitem{Avanthay:2003}
C.~Avanthay, A.~Hertz, and N.~Zufferey.
\newblock A variable neighborhood search for graph coloring.
\newblock {\em European Journal of Operational Research}, 151:379--388, 2003.

\bibitem{Baeck:1996}
T.~B{\"a}ck.
\newblock {\em {E}volutionary Algorithms in Theory and Practice: Evolution
  Strategies, Evolutionary Programming, Genetic Algorithms}.
\newblock Oxford University Press, Oxford, 1996.

\bibitem{Blochliger2003}
I.~Bl{\"o}chliger and N.~Zufferey.
\newblock A reactive tabu search using partial solutions for the graph coloring
  problem.
\newblock In D.~Kral and J.~Sgall, editors, {\em Coloring Graphs from Lists
  with Bounded Size of their Union: Result from Dagstuhl Seminar 03391}, volume
  156 of {\em ITI-Series}, Department of Applied Mathematics and Institute for
  Theoretical Computer Science, Prague, 2003.

\bibitem{Blochliger:2008}
I.~Bl{\"o}chliger and N.~Zufferey.
\newblock A graph coloring heuristic using partial solutions and a reactive
  tabu scheme.
\newblock {\em Computers \& Operations Research}, 35(3):960--975, 2008.

\bibitem{Blum:2011}
C.~Blum, J.~Puchinger, G.A. Raidl, and A.~Roli.
\newblock Hybrid metaheuristics in combinatorial optimization: A survey.
\newblock {\em Applied Soft Computing}, 11(6):4135--4151, 2011.

\bibitem{Blum:2003}
C.~Blum and A.~Roli.
\newblock Metaheuristics in combinatorial optimization: Overview and conceptual
  comparison.
\newblock {\em ACM Computing Surveys}, 35(3):268--308, 2003.

\bibitem{Boettcher:2004}
S.~Boettcher and A.G. Percus.
\newblock Extremal optimization at the phase transition of the three-coloring
  problem.
\newblock {\em Physical Review E}, 69(6):066--073, 2004.

\bibitem{Bondy:2008}
J.A. Bondy and U.S.R. Murty.
\newblock {\em Graph Theory}.
\newblock Springer-Verlag, Berlin, 2008.

\bibitem{Brelaz:1979}
D.~Brelaz.
\newblock {N}ew methods to color vertices of a graph.
\newblock {\em Communications of the ACM}, 22:251--256, 1979.

\bibitem{Brown:1972}
R.~Brown.
\newblock Chromatic scheduling and the chromatic number problem.
\newblock {\em Management Science}, 19(4):456--463, 1972.

\bibitem{Burke:2007}
E.K. Burke, B.~McCollum, A.~Meisels, S.~Petrovic, and R.~Qu.
\newblock A graph-based hyper-heuristic for timetabling problems.
\newblock {\em European Journal of Operational Research}, 176(1):177--192,
  2007.

\bibitem{Chams:1987}
M.~Chams, A.~Hertz, and D.~{de~Werra}.
\newblock Some experiments with simulated annealing for coloring graphs.
\newblock {\em European Journal of Operational Research}, 32:260--266, 1987.

\bibitem{Cheeseman:1991}
P.~Cheeseman, B.~Kanefsky, and W.M. Taylor.
\newblock {W}here the really hard problems are.
\newblock In {\em Proceedings of the International Joint Conference on
  Artificial Intelligence}, volume~1, pages 331--337, San Mateo, 1991. Morgan
  Kaufmann.

\bibitem{website:Chiarandini}
M.~Chiandini and T.~St{\"u}tzle.
\newblock {O}nline compendium to the article: An analysis of heuristics for
  vertex colouring.
\newblock \url{http://www.imada.sdu.dk/~marco/gcp-study/}.
\newblock Accessed 20 December 2010.

\bibitem{Chiarandini:2007}
M.~Chiarandini, I.~Dumitrescu, and T.~St{\"u}tzle.
\newblock Stochastic local search algorithms for the graph colouring problem.
\newblock In T.F. Gonzalez, editor, {\em Handbook of Approximation Algorithms
  and Metaheuristics}, Computer \& Information Science Series, pages
  63.1--63.17. Chapman \& Hall/CRC, Boca Raton, FL, USA, 2007.
\newblock Preliminary version available as \emph{Tech.~Rep.}~AIDA-05-03 at
  Intellectics Group, Computer Science Department, Darmstadt University of
  Technology, Darmstadt, Germany.

\bibitem{Chiarandini:2002}
M.~Chiarandini and T.~St{\"u}tzle.
\newblock An application of iterated local search to graph coloring.
\newblock In D.S. Johnson, A.~Mehrotra, and M.~Trick, editors, {\em Proceedings
  of the Computational Symposium on Graph Coloring and its Generalizations},
  pages 112--125, Ithaca, New York, USA, September 2002.

\bibitem{Chiarandini:2010}
M.~Chiarandini and T.~St{\"u}tzle.
\newblock An analysis of heuristics for vertex colouring.
\newblock In P.~Festa, editor, {\em Proceedings of the 9th International
  Symposium}, volume 6049 of {\em Lecture Notes in Computer Science}, pages
  326--337. Springer, 2010.

\bibitem{Chow:1990}
F.C. Chow and J.L. Hennessy.
\newblock The priority-based coloring approach to register allocation.
\newblock {\em ACM Transactions on Programming Languages and Systems},
  12(4):501--536, 1990.

\bibitem{website:Culberson}
J.~Culberson.
\newblock {G}raph {C}oloring {P}age.
\newblock \url{http://web.cs.ualberta.ca/~joe/Coloring/}.
\newblock Accessed 20 December 2010.

\bibitem{Culberson:1996}
J.~Culberson and F.~Luo.
\newblock {E}xploring the k-colorable landscape with iterated greedy.
\newblock In D.S. Johnson and M.A. Trick, editors, {\em {C}liques, Coloring and
  Satisfiability: {S}econd {DIMACS} {I}mplementation {C}hallenge}, pages
  245--284. American Mathematical Society, Rhode Island, 1996.

\bibitem{deWerra:1985}
D.~{de Werra}.
\newblock An introduction to timetabling.
\newblock {\em European Journal of Operational Research}, 19(2):151--162,
  February 1985.

\bibitem{deWerra:1999}
D.~{de~Werra}, C.~Eisenbeis, S.~Lelait, and B.~Marmol.
\newblock On a graph-theoretical model for cyclic register allocation.
\newblock {\em Discrete Applied Mathematics}, 93(2--3):191--203, 1999.

\bibitem{Demsar:2006}
J.~Dem\v{s}ar.
\newblock Statistical comparisons of classifiers over multiple data sets.
\newblock {\em Journal of Machine Learning Research}, 7:1--30, December 2006.

\bibitem{Dorne:1998}
R.~Dorne and J.K. Hao.
\newblock A new genetic local search algorithm for graph coloring.
\newblock In A.E. Eiben, T.~B{\"a}ck, M.~Schoenauer, and H.P. Schwefel,
  editors, {\em Parallel Problem Solving from Nature - PPSN V, 5th
  International Conference}, volume 1498 of {\em Lecture Notes in Computer
  Science}, pages 745--754. Springer Verlag, Berlin, Germany, 1998.

\bibitem{Eiben:1999}
A.E. Eiben, R.~Hinterding, and Z.~Michalewicz.
\newblock {P}arameter control in evolutionary algorithms.
\newblock {\em IEEE Transactions on Evolutionary Computation}, 3:124--141,
  1999.

\bibitem{Eiben:2003}
A.E. Eiben and J.E. Smith.
\newblock {\em {I}ntroduction to Evolutionary Computing}.
\newblock Springer-Verlag, Berlin, 2003.

\bibitem{Eiben:1998}
A.E. Eiben, J.K. {Van Der Hauw}, and J.I. {Van Hemert}.
\newblock Graph coloring with adaptive evolutionary algorithms.
\newblock {\em Journal of Heuristics}, 4(1):25--46, 1998.

\bibitem{Fleurent:1996}
C.~Fleurent and J.~Ferland.
\newblock Genetic and hybrid algorithms for graph coloring.
\newblock {\em Annals of Operations Research}, 63:437--464, 1996.

\bibitem{Friedman:1937}
M.~Friedman.
\newblock The use of ranks to avoid the assumption of normality implicit in the
  analysis of variance.
\newblock {\em Journal of the American Statistical Association}, 32:675--701,
  December 1937.

\bibitem{Friedman:1940}
M.~Friedman.
\newblock A comparison of alternative tests of significance for the problem of
  m rankings.
\newblock {\em The Annals of Mathematical Statistics}, 11:86--92, March 1940.

\bibitem{Galinier:1999}
P.~Galinier and J.K. Hao.
\newblock Hybrid evolutionary algorithms for graph coloring.
\newblock {\em Journal of Combinatorial Optimization}, 3(4):379--397, 1999.

\bibitem{Galinier:2006}
P.~Galinier and A.~Hertz.
\newblock A survey of local search methods for graph coloring.
\newblock {\em Computers \& Operations Research}, 33:2547--2562, 2006.

\bibitem{Galinier:2008}
P.~Galinier, A.~Hertz, and N.~Zufferey.
\newblock An adaptive memory algorithm for the $k$-coloring problem.
\newblock {\em Discrete Applied Mathematics}, 156(2):267--279, 2008.

\bibitem{Gamache:2007}
M.~Gamache, A.~Hertz, and J.O. Ouellet.
\newblock A graph coloring model for a feasibility problem in monthly crew
  scheduling with preferential bidding.
\newblock {\em Computers \& Operations Research}, 34(8):2384--2395, 2007.

\bibitem{Gamst:1986}
A.~Gamst.
\newblock Some lower bounds for a class of frequency assignment problems.
\newblock {\em IEEE Transactions of Vehicular Technology}, 35:8--14, 1986.

\bibitem{Garey:1979}
M.R. Garey and D.S. Johnson.
\newblock {\em Computers and Intractability: A Guide to the Theory of
  NP-Completeness}.
\newblock W.H. Freeman \& Co., New York, NY, USA, 1979.

\bibitem{Garey:1976}
M.R. Garey, D.S. Johnson, and H.C. So.
\newblock An application of graph coloring to printed circuit testing.
\newblock {\em IEEE Transactions on Circuits and Systems}, 23:591--599, 1976.

\bibitem{Glass:2002}
C.~Glass.
\newblock Bag rationalization for a food manufacturer.
\newblock {\em Journal of the Operational Research Society}, 53:544--551, 2002.

\bibitem{Glover:1986}
F.~Glover.
\newblock Future paths for integer programming and links to artificial
  intelligence.
\newblock {\em Computers {\&} Operations Research}, 13(5):533--549, 1986.

\bibitem{Hamiez:2010}
J.P. Hamiez, J.K. Hao, and F.~Glover.
\newblock A study of tabu search for coloring random 3-colorable graphs around
  the phase transition.
\newblock Technical report, LERIA, Université d'Angers, December 2010.

\bibitem{Hayes:2003}
B.~Hayes.
\newblock {O}n the threshold.
\newblock {\em American Scientist}, 91:12--17, 2003.

\bibitem{Hertz:1987}
A.~Hertz and D.~{de~Werra}.
\newblock Using tabu search techniques for graph coloring.
\newblock {\em Computing}, 39:345--351, December 1987.

\bibitem{Hertz:2008}
A.~Hertz, M.~Plumettaz, and N.~Zufferey.
\newblock Variable space search for graph coloring.
\newblock {\em Discrete Applied Mathematics}, 156(13):2551--2560, 2008.

\bibitem{Hoos:2005}
H.H. Hoos and T.~St{\"u}tzle.
\newblock {\em Stochastic Local Search: Foundations and Applications}.
\newblock Morgan Kaufmann, San Francisco, USA, 2005.

\bibitem{Igel:2003}
C.~Igel and M.~Toussaint.
\newblock {N}eutrality and self-adaptation.
\newblock {\em Natural Computing: An International Journal}, 2:117--132, 2003.

\bibitem{Johnson:1991}
D.S. Johnson, C.R. Aragon, L.A. McGeoch, and C.~Schevon.
\newblock Optimization by simulated annealing: An experimental evaluation,
  {P}art {II}; {G}raph coloring and number partitioning.
\newblock {\em Operations Research}, 39(3):378--406, 1991.

\bibitem{Johnson:1996}
D.S. Johnson and M.A. Trick.
\newblock {\em Cliques, Coloring, and Satisfiability: Second DIMACS
  Implementation Challenge}, volume~26.
\newblock American Mathematical Society, Providence, USA, 1996.

\bibitem{Kimura:1968}
M.~Kimura.
\newblock {E}volutionary rate at the molecular level.
\newblock {\em Nature}, 217:624--626, 1968.

\bibitem{Kubale:2004}
M.~Kubale.
\newblock {\em {G}raph Colorings}.
\newblock American Mathematical Society, Rhode Island, 2004.

\bibitem{Leighton:1979}
F.T. Leighton.
\newblock A graph coloring algorithm for large scheduling problems.
\newblock {\em Journal of Research of the National Bureau of Standards},
  84(6):489--506, 1979.

\bibitem{Lu:2010}
Z.~L{\"u} and J.K. Hao.
\newblock A memetic algorithm for graph coloring.
\newblock {\em European Journal of Operational Research}, 203:241--250, 2010.

\bibitem{Mabed:2010}
H.~Mabed, A.~Caminada, and J.K. Hao.
\newblock Genetic tabu search for robust fixed channel assignment under dynamic
  traffic data.
\newblock {\em Computational Optimization and Applications}, pages 1--24, 2010.
\newblock 10.1007/s10589-010-9376-9.

\bibitem{Malaguti:2008}
E.~Malaguti, M.~Monaci, and P.~Toth.
\newblock A metaheuristic approach for the vertex coloring problem.
\newblock {\em INFORMS Journal on Computing}, 20:302--316, 2008.

\bibitem{Malaguti:2009}
E.~Malaguti and P.~Toth.
\newblock A survey on vertex coloring problems.
\newblock {\em International Transactions in Operational Research}, pages
  1--34, 2009.

\bibitem{Merz:1999}
P.~Merz and B.~Freisleben.
\newblock {F}itness landscapes and memetic algorithm design.
\newblock In D.~Corne, M.~Dorigo, and F.~Glover, editors, {\em {N}ew Ideas in
  Optimization}, pages 245--260. McGraw-Hill, Cambridge, 1999.

\bibitem{Michalewicz:1992}
Z.~Michalewicz.
\newblock {\em {G}enetic Algorithms + Data Structures = Evolution Programs}.
\newblock Springer-Verlag, Berlin, 1992.

\bibitem{Petford:1989}
A.D. Petford and D.J.A. Welsh.
\newblock {A} randomized 3-coloring algorithm.
\newblock {\em Discrete Mathematic}, 74:253--261, 1989.

\bibitem{Smith:1998}
D.H. Smith, S.~Hurley, and S.U. Thiel.
\newblock Improving heuristics for the frequency assignment problem.
\newblock {\em European Journal of Operational Research}, 107(1):76--86, 1998.

\bibitem{Stadler:1995}
P.~Stadler.
\newblock Towards a theory of landscapes.
\newblock In R.~Lopez-Pena, editor, {\em Complex Systems and Binary Networks},
  volume 461 of {\em Lecture Notes in Physics}, pages 77--163.
  Springer-Verlag, Berlin, 1995.

\bibitem{website:Trick}
M.~Trick.
\newblock Network resources for coloring a graph.
\newblock \url{http://mat.gsia.cmu.edu/COLOR/color.html}.
\newblock Accessed 20 December 2010.

\bibitem{Turner:1988}
J.S. Turner.
\newblock {A}lmost all k-colorable graphs are easy to color.
\newblock {\em Journal of Algorithms}, 9:63--82, 1988.

\bibitem{website:Hemert}
J.I. {Van Hemert}.
\newblock {J}ano's {H}omepage.
\newblock \url{http://www.vanhemert.co.uk/csp-ea.html}.
\newblock Accessed 20 December 2010.

\end{thebibliography}
\end{document}